\documentclass{article}
\usepackage{arxiv}

\usepackage[utf8]{inputenc} %
\usepackage[T1]{fontenc}    %
\usepackage{hyperref}       %
\usepackage{url}            %
\usepackage{booktabs}       %
\usepackage{amsfonts}       %
\usepackage{nicefrac}       %
\usepackage{microtype}      %
\usepackage{lipsum}		%
\usepackage{graphicx}
\usepackage{natbib}
\usepackage{doi}
\usepackage[caption=false,font=normalsize,labelfont=sf,textfont=sf]{subfig}

\usepackage{amsmath}
\usepackage{adjustbox}
\usepackage{booktabs}
\usepackage{framed,multirow}
\usepackage{xcolor}
\newcommand{\caesar}{\textsc{caesar}}
\newcommand{\scorpio}{\textsc{Scorpio}}

\title{RADiff: Controllable Diffusion Models for Radio Astronomical Maps Generation}

\author{ \href{https://orcid.org/0000-0002-3906-797X}{\includegraphics[scale=0.06]{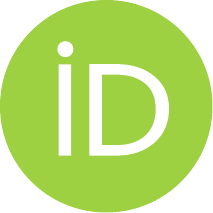}\hspace{1mm}Renato ~Sortino}\\
	PeRCeiVe Lab, INAF\\
	University of Catania\\
	Catania, Italy \\
	\texttt{renato.sortino@phd.unict.it} \\
	\And
	\href{https://orcid.org/0000-0001-6519-5011}{\includegraphics[scale=0.06]{orcid.pdf}\hspace{1mm}Thomas Cecconello} \\
	PeRCeiVe Lab, INAF\\
	University of Catania\\
	Catania, Italy \\
	\And
	\href{https://orcid.org/0000-0001-5921-2372}{\includegraphics[scale=0.06]{orcid.pdf}\hspace{1mm}Andrea DeMarco} \\
    Institute of Space Sciences and Astronomy \\
	University of Malta\\
	Msida, Malta \\
    \And
	\href{https://orcid.org/0000-0001-8687-6609}{\includegraphics[scale=0.06]{orcid.pdf}\hspace{1mm}Giuseppe Fiameni} \\
	NVIDIA AI Technology Center\\
	Milan, Italy \\
    \And
    \href{https://orcid.org/0000-0001-6868-7943}{\includegraphics[scale=0.06]{orcid.pdf}\hspace{1mm}Andrea Pilzer} \\
	NVIDIA AI Technology Center\\
	Milan, Italy \\
	\And
    \href{https://orcid.org/0000-0002-6097-2747}{\includegraphics[scale=0.06]{orcid.pdf}\hspace{1mm}Andrew M. Hopkins} \\
	School of Mathematical and Physical Sciences\\
	Macquarie University\\
	Sydney, Australia \\
	\And
    \href{https://orcid.org/0000-0002-5624-0658}{\includegraphics[scale=0.06]{orcid.pdf}\hspace{1mm}Daniel Magro} \\
    Institute of Space Sciences and Astronomy \\
	University of Malta\\
	Msida, Malta \\
	\And
    \href{https://orcid.org/0000-0001-6368-8330}{\includegraphics[scale=0.06]{orcid.pdf}\hspace{1mm}Simone Riggi} \\
	INAF\\
	Catania, Italy \\
    \And
    \href{https://orcid.org/0000-0002-5574-2787}{\includegraphics[scale=0.06]{orcid.pdf}\hspace{1mm}Eva Sciacca} \\
	INAF\\
	Catania, Italy \\
    \And
    \href{https://orcid.org/0000-0002-3137-473X}{\includegraphics[scale=0.06]{orcid.pdf}\hspace{1mm}Adriano Ingallinera} \\
	INAF\\
	Catania, Italy \\
    \And
    \href{https://orcid.org/0000-0002-7703-0692}{\includegraphics[scale=0.06]{orcid.pdf}\hspace{1mm}Cristobal Bordiu} \\
	INAF\\
	Catania, Italy \\
    \And
    \href{https://orcid.org/0000-0002-3429-2481}{\includegraphics[scale=0.06]{orcid.pdf}\hspace{1mm}Filomena Bufano} \\
	INAF\\
	Catania, Italy \\
    \And
	\href{https://orcid.org/0000-0001-6653-2577}{\includegraphics[scale=0.06]{orcid.pdf}\hspace{1mm}Concetto Spampinato} \\
	PeRCeiVe Lab\\
	University of Catania\\
	Catania, Italy \\
}

\date{}

\renewcommand{\headeright}{}
\renewcommand{\undertitle}{}
\renewcommand{\shorttitle}{RADiff: Controllable Diffusion Models for Radio Astronomical Maps Generation}

\hypersetup{
pdftitle={RADiff: Controllable Diffusion Models for Radio Astronomical Maps Generation},
pdfsubject={cs.AI, cs.CV, physics.space-ph},
pdfauthor={Renato Sortino, Thomas Cecconello, Andrea DeMarco, Giuseppe Fiameni, Andrea Pilzer, Andrew M. Hopkins, Daniel Magro, Simone Riggi, Eva Sciacca, Adriano Ingallinera,  Cristobal Bordiu, Filomena Bufano, Concetto Spampinato},
pdfkeywords={diffusion models, radioastronomy, generative models, semantic image synthesis},
}

\begin{document}
\maketitle

\begin{abstract}
	Along with the nearing completion of the Square Kilometre Array (SKA), comes an increasing demand for accurate and reliable automated solutions to extract valuable information from the vast amount of data it will allow acquiring.
Automated source finding is a particularly important task in this context, as it enables the detection and classification of astronomical objects.
Deep-learning-based object detection and semantic segmentation models have proven to be suitable for this purpose.
However, training such deep networks requires a high volume of labeled data, which is not trivial to obtain in the context of radio astronomy. Since data needs to be manually labeled by experts, this process is not scalable to large dataset sizes, limiting the possibilities of leveraging deep networks to address several tasks.
In this work, we propose RADiff, a generative approach based on conditional diffusion models trained over an annotated radio dataset to generate synthetic images, containing radio sources of different morphologies, to augment existing datasets and reduce the problems caused by class imbalances. We also show that it is possible to generate fully-synthetic image-annotation pairs to automatically augment any annotated dataset.
We evaluate the effectiveness of this approach by training a semantic segmentation model on a real dataset augmented in two ways: 1) using synthetic images obtained from real masks, and 2) generating images from synthetic semantic masks. We show an improvement in performance when applying augmentation, gaining up to 18\% in performance when using real masks and 4\% when augmenting with synthetic masks.
Finally, we employ this model to generate large-scale radio maps with the objective of simulating Data Challenges.
\end{abstract}

\section{Introduction}

The Square Kilometre Array (SKA) will be the largest radio telescope ever built, poised to revolutionize our understanding of the Universe~\citep{dewdney2009square}. With unprecedented levels of sensitivity and spatial resolution, it is expected to facilitate astronomical discoveries across different domains. %
However, the data volume produced by its precursor telescopes already requires considerable effort and well-designed algorithms to extract scientific information in an efficient and automated way~\citep{hotan2021}.

Machine learning, and in particular deep learning, has recently emerged as a valuable tool for detecting and classifying radio sources in images, which is a challenging task, especially for observations with significant diffuse backgrounds or very extended or diffuse sources (i.e., those aiming at the Galactic Plane)~\citep{sortino2023radio,wu2019radio,gheller2018deep,becker2021cnn,lukic2019convosource}.
Background noise and artifacts introduced during the imaging process, such as sidelobes around bright sources, often lead these models to false detections. The identification of sources from multiple non-contiguous islands and their classification into known classes of astrophysical objects poses another challenging task, relevant especially when searching for Galactic objects in Galactic plane surveys.

To achieve high accuracy and precision through deep learning algorithms, a great number of data samples with high-quality annotations are crucial. While SKA data is not yet available, data simulations are commonly employed for testing source extraction tools, as demonstrated in studies such as~\citep{bonaldi2021square,hopkins2015askap,riggi2019caesar,boyce2023hydra}. One of the limitations of the simulated datasets employed in these studies is the assumed ideal background and morphology of extended sources, often modeled as single-component islands generated from a 2D Gaussian distribution of random minor/major axis ratios. Our aim is to overcome this limitation through the application of conditional generative models.\\
Generative models are deep learning architectures that allow for synthesizing new data points that follow the distribution of the training data. Once trained, these models are capable of generating new data points starting from a random Gaussian noise vector.\\
The applicability of generative models to specific tasks and domains stems from the possibility of controlling the output of the generation process. Enforcing constraints on the shape of the generated data and on their features enables the user to generate samples in a specific way to tackle a particular problem~\citep{kucera2022conditional,havaei2021conditional}. For instance, these models can be employed for data augmentation~\citep{antoniou2017data,yin2023ttida} when lack of data availability is an issue or for datasets characterized by heavy class imbalances. In this case, using a controllable generative model can alleviate the problem by allowing for generating under-represented classes and compensating class imbalance~\citep{sampath2021survey}.

While Generative Adversarial Networks (GANs)~\citep{goodfellow2014generative} are widely employed in several domains, including radio astronomy~\citep{glaser2019radiogan,balakrishnan2021pulsar,guo2019pulsar}, we base our method on a more recent generative model architecture: Diffusion Models~\citep{ho2020denoising}. This choice is motivated by the widespread success of this kind of model in the computer vision domain~\citep{ho2020denoising,dhariwal2021diffusion} but, most importantly, by their versatility in terms of controllable generation~\citep{rombach2022high,huang2023composer,zhang2023adding}. An interesting contribution in radio astronomy using conditional diffusion models has recently been made by Wang et al.~\citep{wang2023conditional}, although they propose a different task than ours since they focus on radio interferometric image reconstruction.\\

\begin{figure}[!h]
    \centering
    \includegraphics[width=0.8\textwidth]{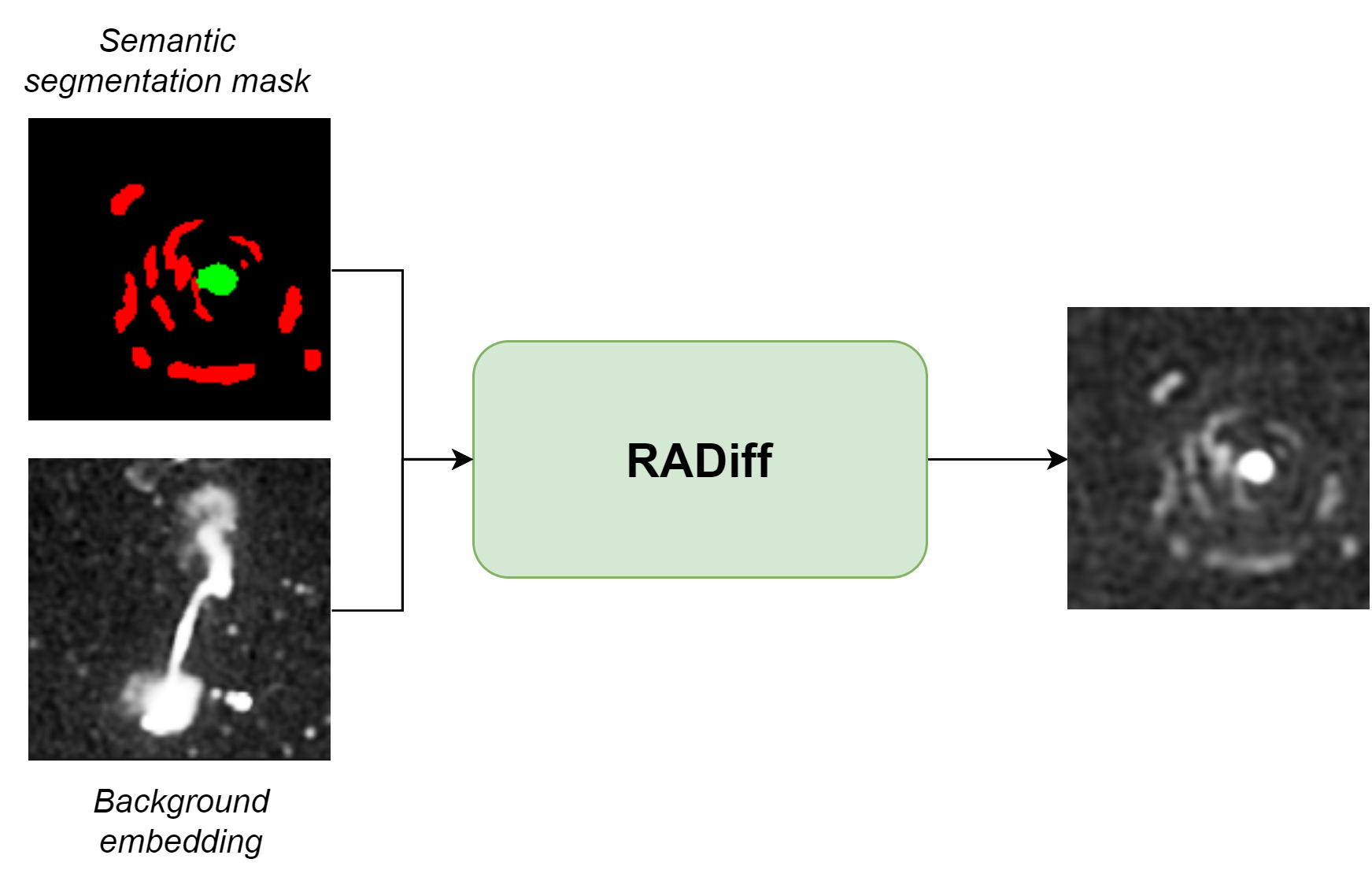}
    \caption{Presentation of the proposed idea. Our model takes as input a segmentation map and an image and generates a synthetic image following the structure defined in the semantic map while transferring the background features of the provided image.}
    \label{fig:teaser}
\end{figure}

We propose \textbf{\emph{RADiff}}, a conditional diffusion model trained to generate radio-astronomical images starting from two types of conditioning: 1) a semantic segmentation map that describes the shape, location, and category of the objects contained in the image to generate, and 2) an image to control the intensity and pattern of the background that surrounds the objects. \\

An overview of the proposed idea is shown in Figure~\ref{fig:teaser}\\

Our key contributions are the following:
\begin{itemize}
    \item We employ a Latent Diffusion Model (LDM)~\citep{rombach2022high} architecture to efficiently generate high-quality images;
    \item We leverage the controllable generation capabilities of LDMs to guide the generation process using a semantic segmentation map and background information from another image;
    \item We evaluate our proposed framework on the domain-specific tasks of data augmentation in cases of insufficient data or class imbalances, and on populating large-scale maps with synthetic radio-astronomical objects.
\end{itemize}
\label{sec:intro}

\section{Related Work}

In this section, we provide an overview of the evolution of generative models in computer vision and their applications to astrophysics and, more specifically, radio astronomy.

\subsection{Generative Models}
Synthetic data generation is a widely explored task involving deep learning models that learn to map a vector sampled from a random Gaussian distribution to data points that follow the distribution of the training data.
This task can be achieved using several neural network architectures. Among these, the most commonly employed are Variational Autoencoders (VAE)~\citep{kingma2013auto}, Generative Adversarial Networks (GAN)~\citep{goodfellow2014generative}, and, from more recent advancements, diffusion models~\citep{ho2020denoising}.
VAEs follow the architecture of autoencoders, composed of an encoder, designed to map the input data to a lower-dimensional latent space, and a decoder, that maps points from the latent space back to the original space.
Since VAEs are generative models, the encoders map the input to a distribution (approximating a Gaussian) of latent vectors instead of a single latent representation. Then, a latent vector is sampled from this distribution and provided to the decoder to generate data.
Variants of this architecture introduce improvements on the reconstruction efficiency by latent space quantization~\citep{van2017neural}, on the controllability of generated samples with class labels~\citep{zhao2017learning}, or on the quality of the generation using the attention mechanism~\citep{lin2019improving}
While VAEs rely on supervised training, GANs~\citep{goodfellow2014generative} employ adversarial training, employing two neural networks, namely a generator and a discriminator, trained in an adversarial way. The former is optimized to generate data that follows the distribution of the training data, while the latter is trained to distinguish between ``real'' data belonging to the training set and samples produced by the generator.
Since these two modules are trained jointly with opposite objective functions, adversarial training can lead to unstable and unpredictable training, often causing the model to diverge. 
Several improvements have been proposed to the original architecture, by adding conditioning information to control the distribution of the generated samples~\citep{mirza2014conditional,isola2017image,reed2016generative}, introducing the ability to map samples back to the latent space from the image space~\citep{karras2020analyzing}, and enabling the possibility of transferring the style of an image to another, without losing its semantic content~\citep{zhu2017unpaired}.
During recent years, Denoising Diffusion Probabilistic Models (DDPM)~\citep{ho2020denoising} have emerged as a valid alternative to adversarial approaches, outperforming them on common benchmarks~\citep{dhariwal2021diffusion}.
These models are trained in a supervised way as a sequence of denoising autoencoders. During training, noise is gradually added to the input data and a neural network learns to revert this process (See Section~\ref{sec:method} for a detailed explanation of how these models work). Diffusion models proved to be flexible in accepting different kinds of conditioning information~\citep{rombach2022high,zhang2023adding} and are also capable of combining multiple inputs for more precise guidance~\citep{huang2023composer}.
We select this family of generative models motivated by their ability to treat multiple conditioning sources other than their improved generation quality with respect to GANs.

\subsection{Generative Models in Astronomy}
Many methods in the radio-astronomical field have successfully shown how generative models, in particular VAEs and GANs, can be successfully employed to serve several purposes.
VAEs can be effectively used to simulate radio galaxies~\citep{bastien2021structured,spindler2021astrovader} for radio surveys enhancing the generation process with conditional information on the class labels.
\citet{Lanusse2021} propose Flow-VAE, a hybrid VAE responsible for learning a representation of the galaxy images, combined with a latent-space normalizing flow. This model is trained to generate the galaxy light profiles, and the outputs can be conditioned on physical galaxy parameters. %
While VAEs can successfully synthesize images in different contexts, GANs show a wider range of applications in radio astronomy. 
\citet{glaser2019radiogan} present RadioGAN, a method capable of performing image-to-image translation between two different radio surveys. It extracts information from radio data and recovers extended flux from a survey with a high angular resolution. This approach can be used for data augmentation and image translation, though the generation process is limited to being controlled by other images and not by other types of information.
\citet{balakrishnan2021pulsar} propose a GAN-based approach to identify new radio pulsars alleviating the problem of lack of labeled data. They propose semi-supervised GANs, which achieve comparable performance to supervised methods using a small fraction of labeled data with respect to unlabeled data. 
GANs have also been employed with the purpose of data augmentation for object detection models to generate realistic solar radio burst (SRB) simulations~\citep{scully2023simulating} and radio galaxies~\citep{kummer2022radio}. In both cases, the generation is not controllable so its applicability to the data augmentation use cases is limited.
Given their recent popularity, diffusion models have been employed in radio astronomy as well~\citep{wang2023conditional}, to perform radio interferometric image reconstruction. This approach uses the original architecture of Diffusion Models~\citep{ho2020denoising} and conditions the generation process on the visibility of the image, to estimate its cleaned version.
To the best of our knowledge, no approach in radio astronomy has explored controllable generative models to perform data augmentation. In particular, we propose a multi-conditional model with two different sources of conditioning information to achieve data augmentation to improve the performance of deep learning models when labeled data is scarce.
\label{sec:related}

\section{Dataset}
\label{sec:dataset}

In this section, we present the dataset (named Survey Collection) employed to train and evaluate our proposed method, along with other approaches, for comparison. The dataset consists of cutouts of large maps obtained from several galactic radio astronomical surveys and segmentation masks used as annotations.

\subsection{Survey Collection}

We train and evaluate our generative models on a dataset, which we name Survey Collection (SC), composed of maps from several galactic radio astronomical surveys. These maps exhibit large sizes (up to 12,000 $\times$ 15,000 pixels), which cannot be directly provided to deep learning models since they would require an excessive amount of computational resources. 
For this reason, we extract cutouts of size $128 \times 128$ from each radio map, making sure that extended objects are not cropped within a cutout. The dataset contains a total of 13,602 map cutouts extracted from several radio astronomical surveys obtained with the following telescopes: 1) the Australia Telescope Compact Array (ATCA), 2) the Australian Square Kilometre Array Pathfinder (ASKAP), and 3) the Very Large Array (VLA). More details on the objects contained in the cutouts are reported in Table~\ref{tab:surveys}.

\begin{table}[!ht]
    \caption{Count of map cutouts split by survey, with details on telescope type and angular resolution. Bmaj and Bmin are expressed in arcsec.}
    \centering
    \begin{adjustbox}{max width=\textwidth}
        \begin{tabular}{lllccc}
            \toprule
            Telescope & Survey & Project & Bmaj & Bmin & Cutouts \\ 
            \midrule
            ATCA & SCORPIO & SCORPIO~\citep{Umana2015} & 9.8  & 5.8  & 2,115 \\
            \midrule
            VLA & FIRST~\citep{becker1995first} & RGZ~\citep{Banfield2015} &  5  & 5  & 3,504 \\
            \midrule
            \multirow{4}*{ASKAP} & \multirow{4}*{EMU} & Early Science, SCORPIO-36~\citep{ingallinera2022evolutionary} & 9.4  & 7.7  & 217 \\
             &  & Early Science, SCORPIO-15~\citep{umana2021first} & 24  & 12  & 3,532  \\
             &  & Pilot 1~\citep{Norris2021} & 12.5  & 10.9  & 351 \\
             &  & Pilot 2 & 16.6  & 13.4  & 3,883 \\
            \midrule
            Total & & & & & 13,602 \\
            \bottomrule
        \end{tabular}
        \end{adjustbox}
    \label{tab:surveys}
\end{table}

We extracted images from the following surveys:
\begin{itemize}
    \item Data Release 1 (DR1)\footnote{\url{https://cloudstor.aarnet.edu.au/plus/s/agKNekOJK87hOh0} from \url{https://github.com/chenwuperth/rgz_rcnn/issues/10}}~\citep{wu2019radio} of the Radio Galaxy Zoo (RGZ) project~\citep{Banfield2015}, using 1.4 GHz radio observations at 5" resolution from the Faint Images of the Radio Sky at Twenty cm (FIRST) survey~\citep{becker1995first};
    \item 1.2 GHz ASKAP-36 \scorpio{} survey at $\sim$9.4"$\times$7.7" resolution (see~\citep{ingallinera2022evolutionary} for a description of the survey).
    \item ASKAP EMU pilot survey at $\sim$12.5"$\times$10.9" resolution~\citep{Norris2021};
    \item 912 MHz ASKAP-15 \scorpio{} survey at $\sim$24"$\times$21" resolution~\citep{umana2021first}
    \item 2.1 GHz ATCA \scorpio{} survey at $\sim$9.8"$\times$5.8" resolution~\citep{Umana2015};
    \item ASKAP EMU Pilot 2 maps at $\sim$16.6" x 13.4" angular resolution observed in Stokes-V polarization%
    
\end{itemize}

\begin{figure}[!h]
    \centering
    \subfloat[ATCA]
        {\includegraphics[width=0.33\textwidth]{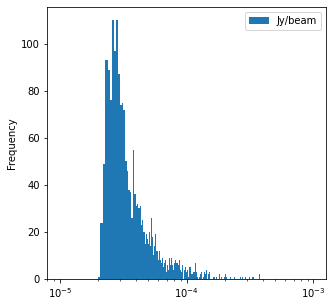}
        \label{subfig:atca}}
    \subfloat[ASKAP]
        {\includegraphics[width=0.33\textwidth]{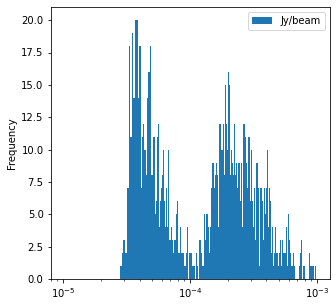}
        \label{subfig:askap}}
    \subfloat[VLA]
        {\includegraphics[width=0.33\textwidth]{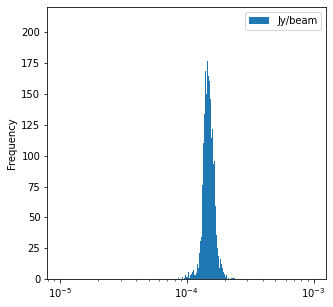}
        \label{subfig:vla}}
    \vspace{2mm}
  \caption{RMS value distribution of images in the dataset for each telescope. Values on the x-axis are expressed in Jy/beam}
  \label{fig:rms}
\end{figure}

In Figure~\ref{fig:rms}, for each telescope we report the RMS values distribution, expressed in Jy/beam. A similar version of this dataset, not containing the Stokes-V radio background maps from ASKAP EMU Pilot 2, has been already used in previous works~\citep{sortino2023radio,riggi2023astronomical}.

Each cutout contains one or multiple objects belonging to one of the following classes:

\begin{itemize}
    \item \emph{Extended Sources}: This object category includes radio galaxies from the RGZ and ASKAP-36 surveys, described above.
    These maps contain extended sources with a 2- and 3-component morphology. 
    Objects from these surveys are extended emission sources and multi-island sources with two, three, or more components. We removed HII regions, planetary nebulae, and supernova remnants, thus all the sources are likely extragalactic.

    \item \emph{Compact sources}: This category contains compact radio sources with single-island morphology. We obtained images with such objects from the ASKAP-15 and ATCA \scorpio{} surveys, reported above.
    
    \item \emph{Spurious sources}: Structures that appear bright but are not categorizable as proper sources fall into this category. This includes mostly imaging artifacts created around bright radio sources due to the limitation of the acquisition instrument. These objects were obtained from the ASKAP EMU pilot survey, ASKAP-15 \scorpio{}, and ASKAP-36 \scorpio{}, ATCA \scorpio{} survey. Traditional algorithms tend to erroneously classify these artifacts as real sources.
     \item \emph{Background}: These images are extracted from the ASKAP EMU Pilot 2 Stokes-V images. %
\end{itemize}

\begin{figure}[t]
    \centering
      \centering
      \includegraphics[width=0.8\linewidth]{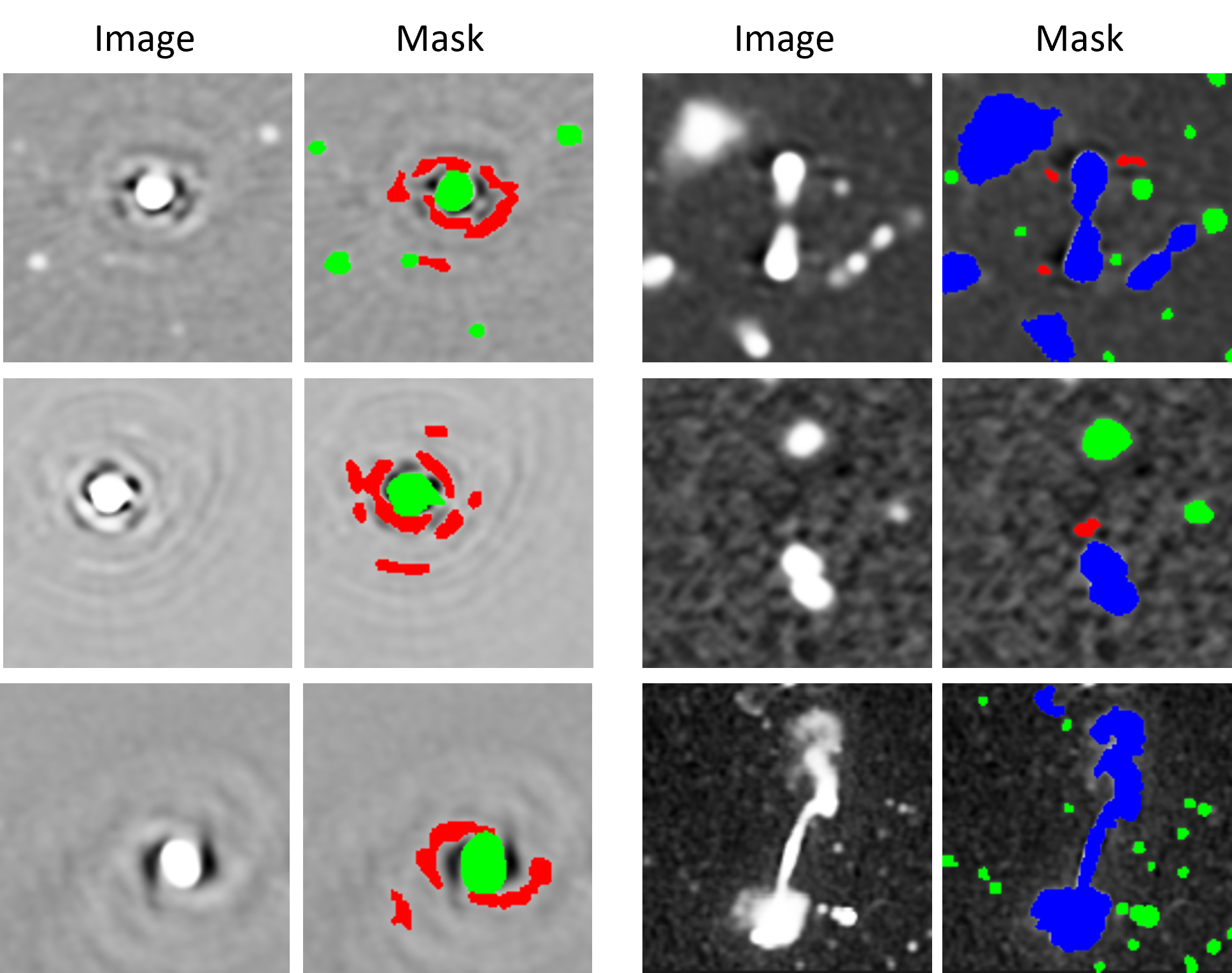}
    \caption{Samples of images and masks used in our dataset. We indicate compact sources in green, extended sources in blue (without distinguishing between single- or multi-island), and spurious sources in red.}
    \label{fig:samples}
\end{figure}

Throughout the paper, we will omit `source' when it becomes redundant and only use their categorization, i.e. one among \emph{`compact'}, \emph{`extended'}, and \emph{`spurious'} and refer to radio map cutouts as images.

The whole dataset, aggregating images from the aforementioned surveys, consists of a total of $36,773$ objects. More details are shown in Table \ref{tab:objects}.

\begin{table}
    \centering
    \caption{Number of object instances and average number of objects in each cutout for each class.}
    \begin{tabular}{ccc}
        \toprule
         \textbf{Object class} &  
         \textbf{\# of objects} &
         \textbf{avg. objects} \\
         & & \textbf{per cutout}
         \\
         \midrule
         Compact & 26,368 & 2.83 \\
         Extended & 4,212 & 1.03 \\
         Spurious & 2,310 & 3.76 \\
          \midrule
         \textit{Total} & 36,773 &  2.7  \\
         \bottomrule
    \end{tabular}
    \label{tab:objects}
\end{table}

Common computer vision approaches process images in common formats (e.g., JPG or PNG) and normalize them to values in the range $\left[0,1 \right]$, so one possible solution to use FITS data would be to convert it into a common image format before feeding it to any model. The problem with this approach is that it involves mapping all (floating point) values to integers in the range $\left[0, 255 \right]$. This contributes to losing the fine-grained details stored in FITS data and altering the dynamic range of the data.
Alternatively, we apply the following pre-processing steps to each FITS image: we first set the NaN values to 0 to avoid problems when giving the images to the model, then transform the data using a tanh function to map the image to the range $\left[0,1 \right]$ and finally normalize the image to avoid unstable training. 

\subsection{Segmentation Masks}
Conditioning a generative model requires additional information on the data samples to be generated so that, at inference time, this information can be used to control the generation process. 
For this reason, we annotate each cutout with a segmentation mask, typically employed in semantic segmentation tasks~\citep{ronneberger2015unet,pino2021semantic,xie2021segformer}.
A segmentation mask is a 2D map with the same size as the image it annotates, where each pixel is an integer describing the category to which that pixel belongs. Given a semantic map $M \in \mathbb{R}^{H \times W}$, and $N_c$ possible classes, each pixel $M(i)$, with $i \in [0, H \times W]$ is defined as follows: 
\begin{equation}
    M(i) = n \in [0, N_c] 
\end{equation}
This way it is possible to determine the shape and number of the objects contained in the cutouts, as well as their category.
We use these annotations to train a conditional generative model, which can be employed to generate realistic radio map cutouts, provided with a segmentation mask. We corroborate the choice of using segmentation maps as conditioning by evaluating our model on the use case of data augmentation for semantic segmentation models (See Section~\ref{sec:augmentation}).

To this purpose, we use the \caesar{} source finder on each cutout to produce a first raw object segmentation, which is then manually refined by domain experts. This step is crucial to correctly identify sources and distinguish proper sources from spurious ones when they are close together, as source finders often detect them as belonging to the same island.
Images belonging to the RGZ survey have been annotated using both infrared and radio data, while annotations of data originating from other surveys are based only on radio data.
Samples of annotated images are reported in Fig.~\ref{fig:samples}.

Both cutouts and segmentation masks are kept under version control, using the Data Version Control (\textsc{dvc}) framework\footnote{\url{https://dvc.org/}}, and stored separately in two different formats. The cutouts are stored in FITS \footnote{\url{https://fits.gsfc.nasa.gov/fits_documentation.html}} files, while the information of the annotations is contained in JSON files, one for each cutout, where all the individual FITS files containing each object mask are specified (see~\citep{sortino2023radio} for more information on the format).

For proper evaluation, we randomly split our dataset into a train and a test set, following a ratio of 80/20, thus ending up with 10,881 and 2,720 images for the train and test set, respectively. \\
\label{sec:method}

\section{Methodology}
\label{sec:method}

\begin{figure}[!ht]
    \centering
    \includegraphics[width=\textwidth]{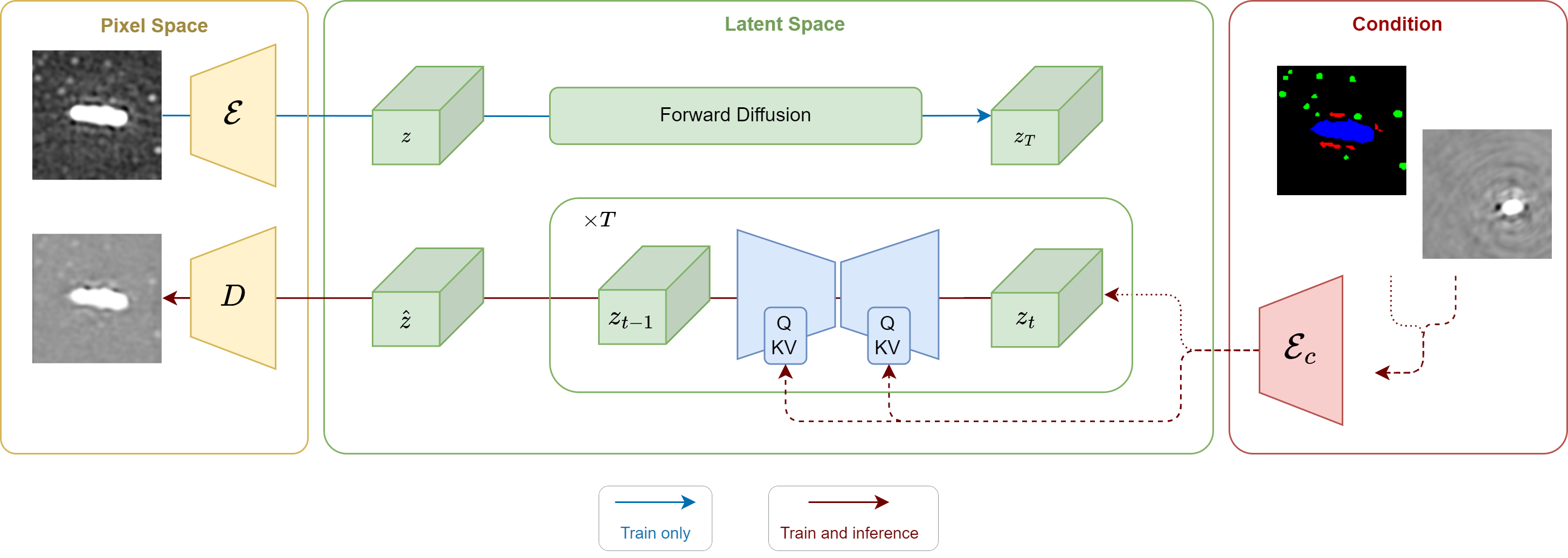}
    \caption{Overview of RADiff, our proposed approach based on diffusion models. The autoencoder projects the data from the pixel space into the latent space $z$ and vice versa. The diffusion model progressively adds noise to the latent vector for a total number $T$ of timesteps and then learns to revert this process, estimating $\hat{z}$. The condition encoder $\mathcal{E}_c$ projects the semantic mask and the background features into a latent embedding to be used by the diffusion model at each timestep $t$. The first one is concatenated to the latent noise vector, while the second one is used in a cross-attention operation with the output of the U-Net intermediate layers.}
    \label{fig:arch}
\end{figure}

In this section, we introduce the architecture of our conditional generative model, named RADiff (Radio Astronomical Diffusion) illustrated in Fig.~\ref{fig:arch}. The model comprises the following key components: 1) an autoencoder ($\mathcal{E}$ and $\mathcal{D}$ in the figure) that projects the images to a latent space, compressing them into a lower-dimensional representation; 2) a diffusion model~\citep{ho2020denoising}, which is an iterative model that represents the core of the generative process; and 3) a conditional encoder $\mathcal{E}_c$, that projects the conditioning information on the latent space to embed it into the generation process.

Diffusion models do not scale well for image sizes above 32 $\times$ 32 or 64 $\times$ 64, due to their multi-head attention operations~\citep{vaswani2017attention}, which involve a complexity of $O(N^2)$.  
In our case, we treat images of size 128 $\times$ 128, so we employ a Latent Diffusion Model (LDM)~\citep{rombach2022high}, which operates in a latent space learned from the dataset in a preliminary phase. In particular, we first train an autoencoder to reconstruct the images in the dataset and then use its learned latent representations as input to the diffusion model. The purpose of using a latent representation instead of full-sized data is the reduced computational complexity needed to perform the operations since such representations allow for compressing the information in the data into smaller feature maps.

One interesting property of LDMs is that they allow for customized generation by supporting multiple conditioning mechanisms~\citep{rombach2022high,huang2023composer}. The generation process can be thus enriched with conditional information to efficiently control the sampled data using human-readable information. We use semantic segmentation maps and image embeddings to condition our diffusion model (see Section~\ref{sec:conditioning}).

\subsection{Learning latent representations}
\label{sec:autoencoder}
As pointed out at the beginning of this section, the choice of employing latent diffusion models implies the necessity of learning a latent vector space from all the samples in the dataset as it will be used as input to the diffusion model.
For this reason, we train an autoencoder~\citep{hinton2006reducing} ($\mathcal{E}$ and $\mathcal{D}$ in Figure~\ref{fig:arch}), to reconstruct the input data, and regularize its learned latent space using a Kullback-Leibler divergence to reduce its variance.
Formally, given an input image $x \in \mathbb{R}^{H \times W \times C}$, with $H, W$ being the image size and $C$ the number of channels, the encoder $\mathcal{E}$ projects it into a latent vector $z = \mathcal{E} \left(x \right) \in \mathbb{R}^{h \times w \times c_z}$ where $h = \frac{H}{f}$, $w = \frac{W}{f}$, $f$ is the \emph{compression factor}, and $c_z$ is the number of channels of the latent vectors (generally, $c_z >> C$). The decoder $\mathcal{D}$ maps the latent vector $z$ back onto the pixel space and produces the reconstructed image $\hat{x} = \mathcal{D}\left(z\right)$.
We train this autoencoder with a reconstruction loss $\mathcal{L}_{rec}$ and a regularization loss $\mathcal{L}_{reg}$. The reconstruction loss, defined in~\ref{eq:reconstruction}, minimizes the distance between the pixels of the reconstructed image w.r.t the input, using a mean absolute error loss.
\begin{equation}
    \mathcal{L}_{rec} = \left \lVert x - \hat{x} \right \rVert_1
    \label{eq:reconstruction}
\end{equation}
The regularization loss, defined in~\ref{eq:regularization}, limits the variance of the latent space, preventing it from growing indefinitely. We use a KL-divergence between the distribution of the latent vectors $P\left(z\right)$ and a standard Gaussian distribution $Q\left(z\right) = \mathcal{N}\left(0,1\right)$.

\begin{equation}
    \mathcal{L}_{reg} = KL(P||Q)=\sum_{z}P(z)\log(\frac{P(z)}{Q(z)})
    \label{eq:regularization}
\end{equation}

The architecture of the autoencoder comprises two downsampling blocks, one bottleneck block, and two upsampling blocks. The downsampling block is made of two residual blocks (ResBlock), one attention block (AttnBlock), and a downsampling convolution. The ResBlock contains a series of 3 convolutions with residual connections~\citep{he2014deep}, the AttnBlock employs multi-head self-attention~\citep{vaswani2017attention} on its input, and the downsampling convolution reduces the spatial size of the feature map while expanding the number of channels. The bottleneck is a sequence of two ResBlocks interleaved with an AttnBlock. The upsampling block is the dual of the downsampling one, with an upsampling convolution at the end, increasing the spatial size of the feature map and reducing its number of channels.

The main shape of the objects is correctly restored, although the highest difference lies at the edges of the objects, as the model loses some high-frequency details due to compression.

\begin{figure}[ht]
    \centering
    \includegraphics[width=\textwidth]{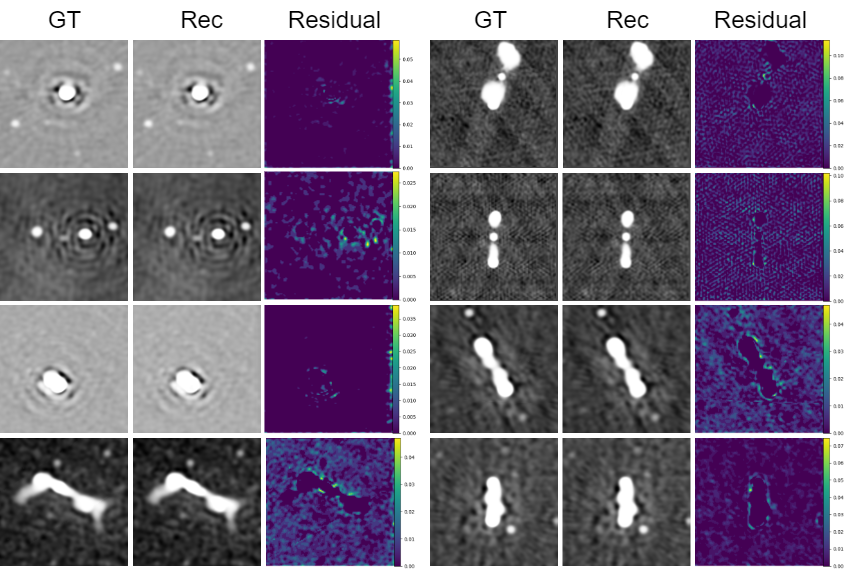}
    \caption{Reconstruction quality of the autoencoder. The ``Residual'' column highlights the difference between ground truth and reconstructed}
    \label{fig:ae}
\end{figure}

\subsection{Diffusion model}
\label{sec:diffusion_model}

The diffusion model~\citep{ho2020denoising} is a probabilistic model that learns to revert a Markov chain of length $T$ with the objective of generating images from Gaussian noise.

\begin{figure}[ht]
    \centering
    \includegraphics[width=\textwidth]{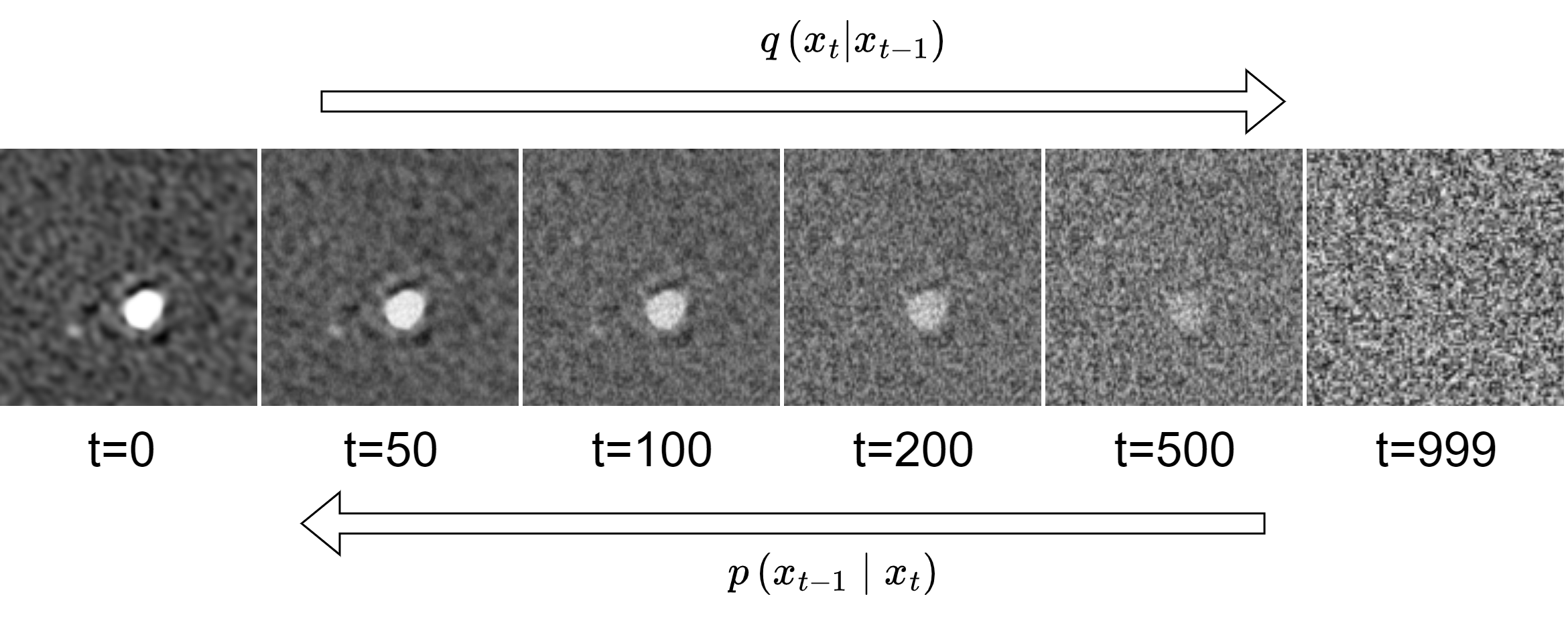}
    \caption{Example of the forward diffusion process. The image is gradually perturbed up to the point of becoming an isotropic Gaussian distribution.}
    \label{fig:diffusion}
\end{figure}

Training a diffusion model consists of two stages: 1) the forward diffusion process, depicted in Figure~\ref{fig:diffusion}, and 2) the backward diffusion process.
During the first stage, data is converted into an isotropic Gaussian distribution by progressively adding Gaussian noise for $T$ timesteps.
Given a data point $x$, sampled from the dataset, we gradually destroy the information it contains by adding noise, following an iterative deterministic process $q \left(x_t | x_{t-1} \right)$, with $t \in \left[ 0, T \right]$ to obtain the noised data. In this process, no neural network is involved as the parameters of the Gaussian noise added at each step are defined beforehand. In particular, we define a parameter, $\beta_t = 1 - \alpha_t$, with $\alpha_t \in [0,1]$, that controls the mean and variance of the noise. $\beta_t$ depends on the timestep $t$ and follows a predefined schedule. Following the original work of DDPM~\citep{ho2020denoising}, we select a linear schedule for $\beta_t$. Equation~\ref{eq:fwd-diff} formally defines the forward diffusion process.
\begin{equation}
    \begin{aligned}
        q\left(x_1, \ldots, x_T \mid x_0\right) & :=\prod_{t=1}^T q\left(x_t \mid x_{t-1}\right) \\
        q\left(x_t \mid x_{t-1}\right) & :=\mathcal{N}\left(x_t ; \sqrt{1-\beta_t} x_{t-1}, \beta_t \mathbf{I}\right)
    \end{aligned}
    \label{eq:fwd-diff}
\end{equation}
The generative capability of diffusion models resides in the backward process, defined in Equation~\ref{eq:back-diff}, where the goal is to revert the forward process to obtain the original, denoised version of the data. 
This process is similar to other generative models (e.g., GANs) as it is the equivalent of mapping a Gaussian noise vector to data. To revert the forward process, we need to compute the reverse conditional probability $q \left(x_{t-1} | x_t \right)$. This would require a calculation over the entire dataset to calculate the prior, which is computationally unfeasible, so we optimize the parameters $\theta$ of a neural network to approximate the distribution $q$ with another distribution, $p$, defined as follows:
\begin{equation}
    p\left(x_{t-1} \mid x_t\right):=\mathcal{N}\left(x_{t-1} ; \mu_\theta\left(x_t, t\right), \Sigma_\theta\left(x_t, t\right)\right),
    \label{eq:back-diff}
\end{equation}
where $\mu_\theta$ and $\Sigma_\theta$ are the mean and variance, respectively, of the noise predicted by the network at timestep $t$.
Reconstructing the data at each timestep $t$ happens by subtracting the estimated noise from the noise vector $x_t$ to obtain $x_{t-1}$.
As we treat 2D data, we employ a U-Net~\citep{ronneberger2015unet}, as done in other works~\citep{ho2020denoising,rombach2022high}. This model follows the original architecture proposed in Ronneberger et al.~\citep{ronneberger2015unet}, comprising a series of blocks of convolutional layers that first downsample the input to a lower-dimensional space to extract low-level features. Then, a bottleneck processes the downsampled output before the latent vector is expanded again through a series of transposed convolutional layers that upsample the data back to the original image size. Equal-sized blocks in the downsample and upsample paths are connected via skip-connections, as in the original architecture, and allow the deeper blocks to take into account low-level features as well. The convolutional blocks are enriched with residual connections and multi-head self-attention blocks to better capture spatial correlations within the latent vectors.
Multi-head attention~\citep{vaswani2017attention} is defined as follows:

\begin{equation}
    \text{Attention}(Q, K, V) = \text{softmax}\left(\frac{Q K^T}{\sqrt{d_k}} \right) V,
    \label{eq:attn}
\end{equation}

where $Q = W_{Q} \cdot \varphi_{i}(x_{t}), K = W_{K} \cdot \varphi_{\theta}(y), V = W_{V} \cdot \varphi_{\theta}(y)$. 

Here, $W_{Q}, W_{K}, W_{V} \in \mathbb{R}^{d \times d_{z}^{i}}$ are learnable projection matrices, $d_{z}^{i}$ represents the number of feature maps at the output of the $i$-th layer, d represents the size of the flattened feature maps, and  $\varphi_{i}(z_{t})$ is the output of the $i$-th U-Net intermediate layer.

We train our diffusion model using a $L_2$ loss between the noise the U-Net model estimates at a specific timestep $\epsilon_\theta\left(x_t, t\right)$ and the noise added during the forward process at the same timestep $\epsilon_{t}$. This loss $\mathcal{L_\text{diff}}$ is defined as follows:

\begin{equation}
    \mathcal{L_\text{diff}}=E_{t, x_0, \epsilon}\left[\left\|\epsilon_{t}-\epsilon_\theta\left(x_t, t\right)\right\|^2\right]
    \label{eq:loss}
\end{equation}

\subsection{Conditioning mechanism}
\label{sec:conditioning}

Providing noise as the only input to the diffusion model enables the generation of samples that follow the distribution of the training dataset. For practical use cases, generative models need to be controllable with additional information to effectively guide them toward generating a specific output.
Diffusion models are particularly flexible for conditional generation. Given any kind of conditioning information $c$, diffusion models can learn the joint distribution $p\left(x_{t-1} \mid x_t, c\right)$, enabling controllable generation on task-specific information. In our case, given the availability of our annotations, we condition our model on semantic segmentation masks, each of which defines the shape, location, and class of the objects in the image (see Section~\ref{sec:dataset}).

We embed this information in our diffusion model by concatenating the segmentation map with the input of each denoising step. This contributes to injecting the spatial correlation of the segmentation map into the generative process to guide the sampled data using the structure of the semantic maps.

Since the semantic map only guides the disposition of the objects without any information on the background, we add another type of conditioning by extracting global information on the image background via another condition encoder. This allows for also controlling the background of the image by providing an image with the desired background (see Section~\ref{sec:cond-results}).

In the conditional configuration, the model optimizes the following loss function, which is an extension of Equation~\ref{eq:loss} that considers the joint probability with the condition $c$:

\begin{equation}
    \mathcal{L_\text{diff}}=E_{t, x_0, \epsilon, c}\left[\left\|\epsilon_{t}-\epsilon_\theta\left(x_t, t, c\right)\right\|^2\right]
\end{equation}
\label{sec:results}

\section{Results}
In this section, we conduct a series of analyses on our proposed pipeline with a threefold objective: 1) assess the quality of the generated samples both quantitatively and qualitatively; 2) evaluate how this synthetic data can be used for data augmentation in semantic segmentation models with few or imbalanced training data; 3) employ our model to produce large-scale maps with a real background for Data Challenge purposes.  

\subsection{Generation Quality}
We explore the generative performance of our proposed approach on the following tasks: unconditional image generation and semantic image synthesis. We report quantitative results on the fidelity of the produced samples with respect to the images in the dataset, and samples of the generated data to provide visual feedback as well. 

\subsubsection{Unconditional Generation}
As a first step, we assess the ability of our model to generate high-quality images that exhibit realism and relevance in the field of radio astronomy. \\
To gauge this aspect, we employ the Fréchet Inception Distance (FID)~\citep{heusel2017gans}, a widely-used metric introduced for GANs evaluation and used more in general for generative models. The FID, defined in Equation~\ref{eq:fid}, expresses the distance between two multivariate normal distributions. We can use this metric to compute the distance between the distribution of the feature vectors extracted from real images and the ones relative to synthetic samples. These feature vectors are typically extracted using the Inception-v3~\citep{szegedy2016rethinking} classifier trained on the ImageNet~\citep{deng2009imagenet} dataset, which contains images of objects and scenes in common contexts. Since these images exhibit different features with respect to our data, we train a self-supervised model on our dataset to extract domain-specific features and obtain a more meaningful comparison, as done in other approaches~\citep{morozov2021self}.
In particular, we train BYOL~\citep{grill2020bootstrap} with a ResNet18~\citep{he2014deep} backbone on our dataset. Once the model has been trained, we employ its backbone to extract a feature vector, of size 512, from each image in the dataset and from the generated samples. Then, we compute the means $\mu_R$, $\mu_G$ and the covariance matrices $\Sigma_R$, $\Sigma_G$ for the feature vectors extracted from the real dataset (R), and from the generated samples (G), respectively. Finally, the computation of the FID metric is defined as follows:
\begin{equation}
    FID = |\mu_R - \mu_G| + tr(\Sigma_R + \Sigma_G - 2(\Sigma_R \Sigma_G)^{\frac{1}{2}})
    \label{eq:fid}
\end{equation}
Here, $tr$ stands for the trace operator in linear algebra.
As this metric represents a distance, a lower score indicates a higher similarity between the features, thus a better quality of the generated images.\\
Table~\ref{tab:res_uncond} shows the performance of several generative architectures trained on our dataset. Since, to the best of our knowledge, we did not find any open-source generative models in radio astronomy, we evaluate the performance of our model against two state-of-the-art unconditional GANs, namely DCGAN~\citep{radford2015unsupervised} and Progressive Growing GAN (PG-GAN)~\citep{karras2017progressive}. We also evaluate DDPM~\citep{ho2020denoising}, which follows the original implementation of diffusion models, to assess the advantages of using a latent diffusion model (LDM). Since training DDPM on images at 128 $\times$ 128 is not tractable for its high computational requirements (See Section~\ref{sec:autoencoder}), we resize the images to 64 $\times$ 64 pixels in this case. The results show how our model outperforms the GAN-based models and DDPM in terms of FID.

\begin{table}
    \caption{Results on unconditional generation using different architectures in terms of FID score (the lower, the better). Best results are highlighted in bold}
    \centering
        {\begin{tabular}{lccc}
            \toprule
            Model & FID $(\downarrow)$ \\ 
            \midrule
            DCGAN~\citep{radford2015unsupervised} & 101.31           \\
            PG-GAN~\citep{karras2017progressive} & 79.33           \\
            DDPM~\citep{ho2020denoising} & 83.26           \\
            \textbf{RADiff (Ours)} & \textbf{33.69}     \\
            \bottomrule
        \end{tabular}
        }  
    \label{tab:res_uncond}
\end{table}

\subsubsection{Conditional Generation}
\label{sec:cond-results}

Since we are interested in conditioning the generation process as well, in addition to measuring the closeness of the real and generated feature vectors we need to evaluate how the synthetic image is coherent with the conditioning input. In our case, we use semantic segmentation masks as conditions, so we are interested in whether the synthetic image contains the objects defined in the mask. For this reason, we define a \emph{Segmentation Score}, which measures the amount of overlap between the input mask and the mask predicted on the generated image using a pre-trained semantic segmentation model. In particular, we measure the Intersection over Union (IoU), defined in~\ref{eq:iou} between the ground truth mask and the estimated segmentation mask. The IoU is typically employed in semantic segmentation tasks~\citep{zhang2022resnest,xie2021segformer,cheng2020panoptic,bao2021beit} and defines the ratio between the correctly classified pixels for each class $k$ (intersection) and the total amount of pixels of objects belonging to class $k$ (union). We compute the average IoU for each image over the classes and finally average the results from all images. Formally, the IoU is defined as follows:
\begin{equation}
    IoU = \frac{1}{N} \sum{IoU_{i}}, \;\;\; with \; i \in \left[ 0, N \right]
    \label{eq:iou}
\end{equation}
where $N$ is the number of images in the dataset. For each image, considering a task where the number of possible classes is $K$, the $IoU_i$ is defined as follows:

\begin{equation}
    IoU_{i} = \frac{1}{K} \sum{\frac{intersection_k}{union_k}}, \;\;\; with\ k \in \left[ 0, K \right]
    \label{eq:iou_i}
\end{equation}

\begin{equation}
    intersection_k = \sum_{j}{pred_k \circ gt_k} \\
\end{equation}
\begin{equation}
    union_k = \sum_{j}{pred_k \cup gt_k}
\end{equation}

Finally, the terms $pred_k$ and $gt_k$ are binary masks relative to each pixel $j$ of the segmentation mask and each class $k$, specifying where, either the prediction or the ground truth maps, report segmentations of class $k$. Formally, these are defined as follows:

\begin{equation}
        pred_k =
    \begin{cases}
        1 & \text{if } pred_{j} = k\\
        0 & \text{otherwise}
    \end{cases}
\end{equation}

where $pred_j$ refers to the class predicted for the pixel $j$. $gt_k$ is computed in the same way for the ground truth mask.

In our analysis, we employ Tiramisu~\citep{jegou2017one} as it has been previously trained on a radio-astronomical dataset~\citep{pino2021semantic,sortino2023radio}.

Another metric that quantifies the generation quality is the Structural Similarity Index Measure (SSIM)~\citep{wang2004image}, which evaluates the affinity between the structures of the two images $x$ and $y$, defined in Equation~\ref{eq:ssim}.
\begin{equation}
  SSIM(x,y) = \frac{(2\mu_x\mu_y + C_1) + (2 \sigma _{xy} + C_2)} 
    {(\mu_x^2 + \mu_y^2+C_1) (\sigma_x^2 + \sigma_y^2+C_2)}
  \label{eq:ssim}
\end{equation}
Here, $\mu$ and $\sigma^2$ indicate mean and variance, respectively, $\sigma_{xy}$ is the covariance between $x$ and $y$, $c_1=(K_1 L)^2$ and $c_2=(K_2 L)^2$ are stabilizing constants. $K_1=0.01$, $K_2=0.03$ by default, $L$ represents the dynamic range of the images.
We compare our model against two state-of-the-art semantic image synthesis approaches, SPADE~\citep{park2019semantic} and INADE~\citep{tan2021diverse}, which serve as baselines. Then, to evaluate both the quality of the generation of our model and its capability of transferring the background properties from another image to the synthetic one, we test our model in two configurations: 1) we condition the diffusion model by providing only the semantic masks as input; 2) we use both the semantic masks and the background information from another image as conditioning (See Figure~\ref{fig:arch}).
We report the quantitative results of this analysis in Table~\ref{tab:res_cond}, showing how using a combination of mask and background conditioning contributes to higher performance and a more stable generation.

\begin{table}
    \caption{Results on semantic image synthesis. Best results are highlighted in bold. \emph{no-bg} represents our model without the conditioning on the background.}
    \centering
        {\begin{tabular}{lcccc}
            \toprule
            Model & FID $(\downarrow)$ & SSIM $(\uparrow)$ & Segmentation Score $(\uparrow)$  \\ 
            \midrule
            Real Images & - & - & 68.67\% \\
            \midrule
            SPADE~\citep{park2019semantic} & 73.49 & 26.84\% & 36.19\% \\
            INADE~\citep{tan2021diverse} & 65.78 & 27.53\% &  38.31\% \\
            \textbf{RADiff (Ours)~\emph{(no-bg)}} & 21.66 & 30.87\% & 49.53\% \\
            \textbf{RADiff (Ours)} & \textbf{13.17} & \textbf{39.04\%} & \textbf{60.96}\% \\
            \bottomrule
        \end{tabular}
        }  
    \label{tab:res_cond}
\end{table}

We corroborate our results with visual samples of the synthetic images, which help to better understand how the model behaves under different configurations and to assess the quality of the generation.

\begin{figure}
    \centering
    \includegraphics[width=0.85\textwidth]{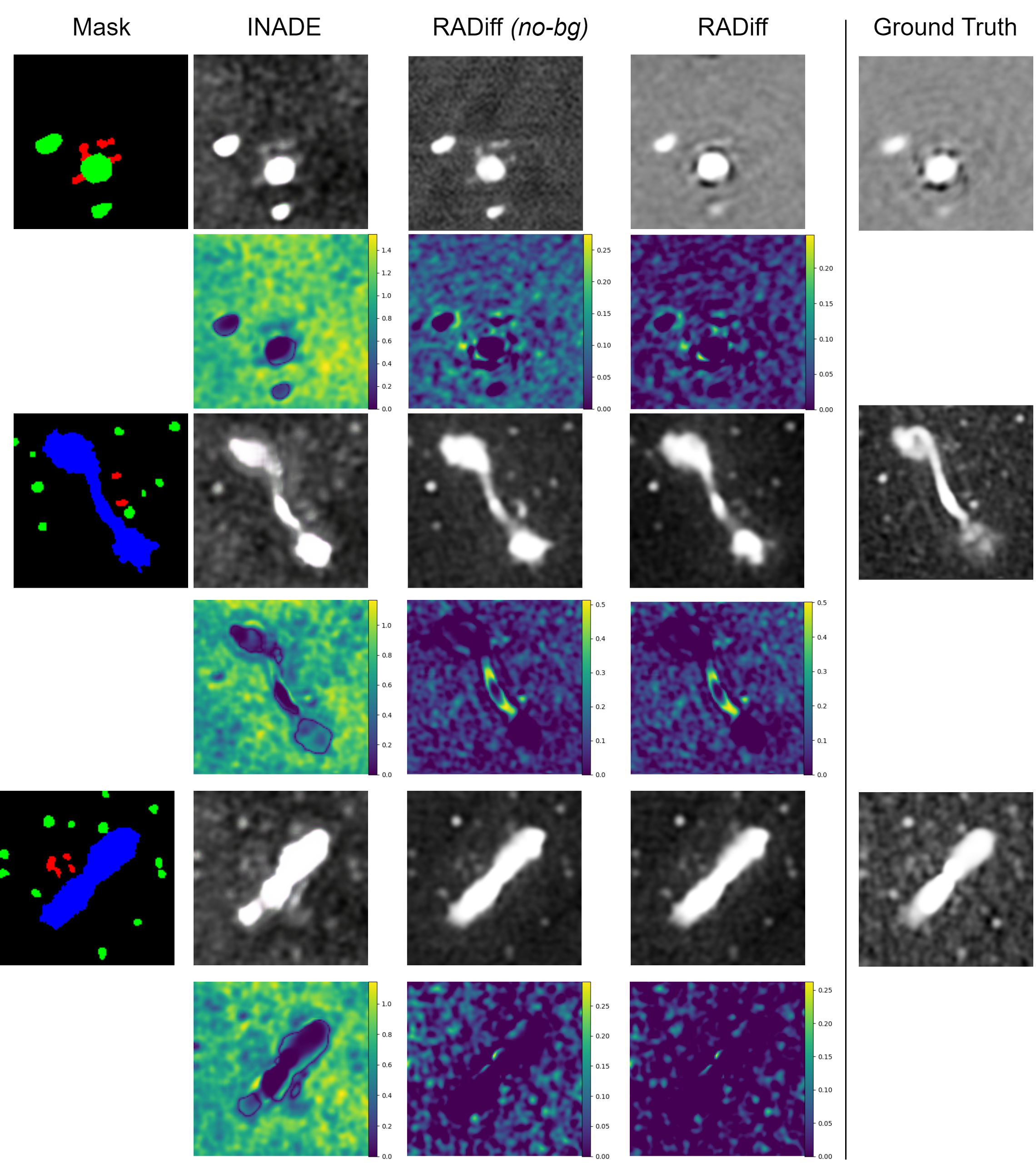}
    \caption{Visual quality of the generated samples in comparison with the input mask and the ground truth images. Each row shows the samples generated from the respective mask, along with the relative ground truth image. Residual images, obtained by subtracting the generated image from the ground truth, are reported below each sample. \textit{no-bg} indicates the generation without conditioning on the background. For the complete RADiff model, we used the ground truth image as background conditioning.}
    \label{fig:generated}
\end{figure}

\begin{figure}
    \centering
    \includegraphics[width=\textwidth]{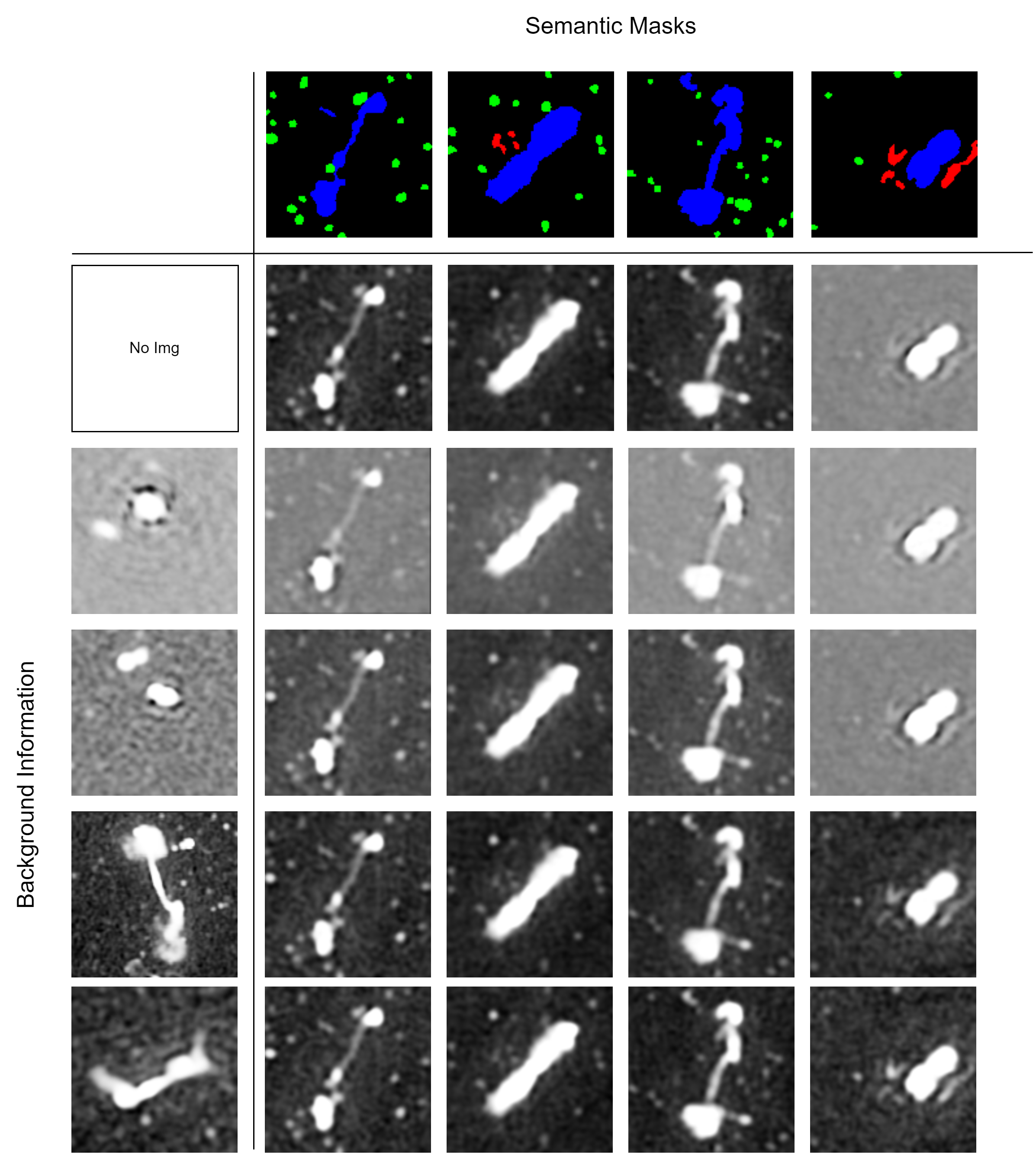}
    \caption{Evaluation of the effect of using different background conditioning on the same masks. All the samples are generated using our complete RADiff model.}
    \label{fig:background}
\end{figure}

In Figure~\ref{fig:generated}, we report some synthetic images obtained with different models, conditioned with different masks. 
While the shape and location of the objects in the input map are correctly reproduced in both cases, the results show that adding background conditioning to the semantic masks reduces the distance from the ground truth, by reproducing the background features with higher fidelity.
The effect of the background conditioning is better visualized in Figure~\ref{fig:background}, where it becomes clear that conditioning the generation process on the background of different images contributes to transferring the visual properties of the background into the synthetic image.

\subsection{Synthetic images for data augmentation}
\label{sec:augmentation}
In contexts where large data volumes are difficult to obtain, generative models offer the possibility to augment available datasets with synthetic images~\citep{antoniou2017data,sandfort2019data}, improving the performance of deep architectures. 
For this use case, we evaluate our approach by training a semantic segmentation model, Tiramisu~\citep{pino2021semantic}, adopting the following strategies to augment our Survey Collection dataset: 1) we generate fully-synthetic image-mask pairs, and 2) we use real semantic masks to generate the synthetic images. We compare the performance of the model instances in terms of IoU, defined in Equation~\ref{eq:iou}. 
In semantic segmentation tasks, the dataset is composed of image-mask pairs, where the image is the input and the mask represents the ground-truth annotation. 
For data augmentation purposes, we generate fully-synthetic image-mask pairs containing compact and extended sources. To generate images using our RADiff pipeline, we need to first produce the segmentation masks to control the image generation and to serve as annotations for training the segmentation model. Ideally, these masks should be designed by experts and contain realistic objects distributions, and shapes. To evaluate our approach, we produce synthetic segmentation masks by training an unconditional diffusion model (DDPM~\citep{ho2020denoising}) to generate 5,000 masks, with a structure similar to those present in our dataset. In Table~\ref{tab:augmented_count}, we report the number of object instances in the generated masks. Then, we use the masks to condition our pre-trained RADiff model and obtain the associated synthetic images. 
In this analysis, we evaluate how synthetic data can solve class imbalances, so we augment the dataset using image-mask pairs containing only objects belonging to a single class among compact and extended and assess their performance on each class separately. For completeness, we also evaluate the model on the dataset augmented on objects from both classes. 
Table~\ref{tab:augmented} shows the results of this analysis and it emerges that augmenting the real dataset with fully-synthetic image-mask pairs can improve the overall performance of the model. In addition to this, controlling the type of objects generated can alleviate performance loss related to class imbalances, as shown by the performance improvement for the augmented class. The approach is limited by the quality of the segmentation masks, which is why, to properly exploit the potential of this method, well-designed masks should be provided to generate the synthetic images.

\begin{table}[!ht]
    \caption{Number of object instances for each class in the generated masks.}
    \centering
        {\begin{tabular}{lccc}
            \toprule
            Class & Augmented & Real & Ratio \\ 
            \midrule
            Compact & 7,851 & 26,368 & 30\% \\
            Extended & 2,134 & 4,212 & 50\% \\
            Spurious & 248 & 2,310 & 10\% \\
            \midrule
            Total & 10,233 & 32,890 & 30\%\\
            \bottomrule
        \end{tabular}
        }  
    \label{tab:augmented_count}
\end{table}

\begin{table}[!ht]
    \caption{Semantic segmentation model (Tiramisu~\citep{pino2021semantic}) trained on the real SC dataset augmented with synthetic image-mask pairs. We first generate 5,000 masks using DDPM~\citep{ho2020denoising} and then feed these masks to RADiff to generate the associated images. The first column represents the objects included in the synthetic masks. All model instances are evaluated on real data, and results are expressed in terms of IoU.}
    \centering
        {\begin{tabular}{lccc}
            \toprule
            Augmentation & All classes & Extended & Compact \\ 
            \midrule
            Compact & 72.29\% & 71.87\% & \textbf{72.87\%} \\
            Extended & \textbf{74.85\%} & \textbf{76.40\%}  & 71.63\% \\
            Both & 72.89\% & 74.51\% & 71.25\% \\
            \midrule
            None & 71.67\% & 72.14\% & 70.72\% \\
            \bottomrule
        \end{tabular}
        }  
    \label{tab:augmented}
\end{table}

Considering these limitations and the fact that synthetic segmentation masks may not represent realistic object distributions and shapes, we evaluate our approach using real segmentation masks to produce synthetic images. To this purpose, we train our RADiff model on a reduced version of our dataset, SC-Reduced ($SC_R$), using a 70\% of the training dataset (7,616 images), then we generate a dataset of augmented synthetic images, SC-Augmented ($SC_A$), providing the semantic masks of the images not belonging to $SC_R$ ($SC_R \displaystyle \cap SC_A = \O$) to RADiff. 
In Table~\ref{tab:datasets_mix}, we report the number of object instances for each category in the two datasets. For a realistic simulation, we left most of the extended sources in the synthetic dataset, since this kind of source is the most difficult to emulate, while compact sources are more easily reproducible.
Then, we train our semantic segmentation model, Tiramisu, in the following configurations: 1) trained only on $SC_R$ to establish a baseline, 2) trained on a mix of $SC_R$ and $SC_A$, and 3) trained on a dataset composed only of synthetic images ($SC_S$), using the real masks from $SC$ to generate them. We evaluate the model on the same test set used for the other experiments and report the results in Table~\ref{tab:data_augmentation}.

\begin{table}[!ht]
    \caption{Object instances after splitting the dataset into the two sub-datasets, $SC_R$, and $SC_A$.}
    \centering
        {\begin{tabular}{lccc}
            \toprule
            Class & $SC_R$ & $SC_A$ \\ 
            \midrule
            Compact & 23,655 & 2,703          \\
            Extended & 1,243 & 2,969          \\
            Spurious & 2,294 & 16          \\
            \midrule
            Total & 27,192 & 5,688          \\
            \bottomrule
        \end{tabular}
        }  
    \label{tab:datasets_mix}
\end{table}

\begin{table}[!ht]
    \caption{We evaluate the impact of extending the dataset with synthetic images conditioned on real masks. We remove 30\% of image-mask pairs from the training set $SC$, obtaining its reduced version $SC_R$, and use these masks to generate a dataset of augmented synthetic images ($SC_A$). $SC_S$ refers to a dataset made of synthetic images obtained from all the masks in the training set of $SC$. Best results are highlighted in bold.}
    \centering
        {\begin{tabular}{lccc}
            \toprule
            Model & All classes & Extended & Compact \\
            \midrule
            $SC_R$ &  63.93\% & 53.02\% & 58.25\% \\
            \hspace{0.1cm} + $SC_A$ & \textbf{70.65}\% &  \textbf{71.64\%} & \textbf{69.52\%} \\
            \midrule
            SC & 71.67\% & 72.14\% & 70.72\% \\
            $SC_S$ & 67.40\%   & 68.68\%  & 58.13\% \\
            \bottomrule
        \end{tabular}
        }  
    \label{tab:data_augmentation}
\end{table}

\subsection{Large-scale map generation}
\label{sec:large-scale}
\begin{figure}[ht]
    \centering
    \subfloat[]
        {\includegraphics[width=0.33\textwidth]{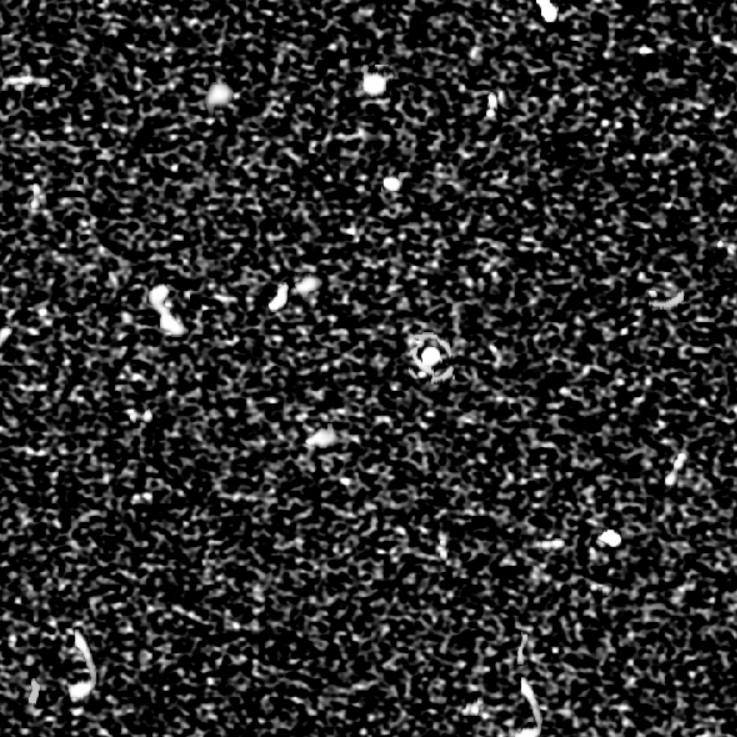}
        \label{subfig:large1}}
    \subfloat[]
        {\includegraphics[width=0.33\textwidth]{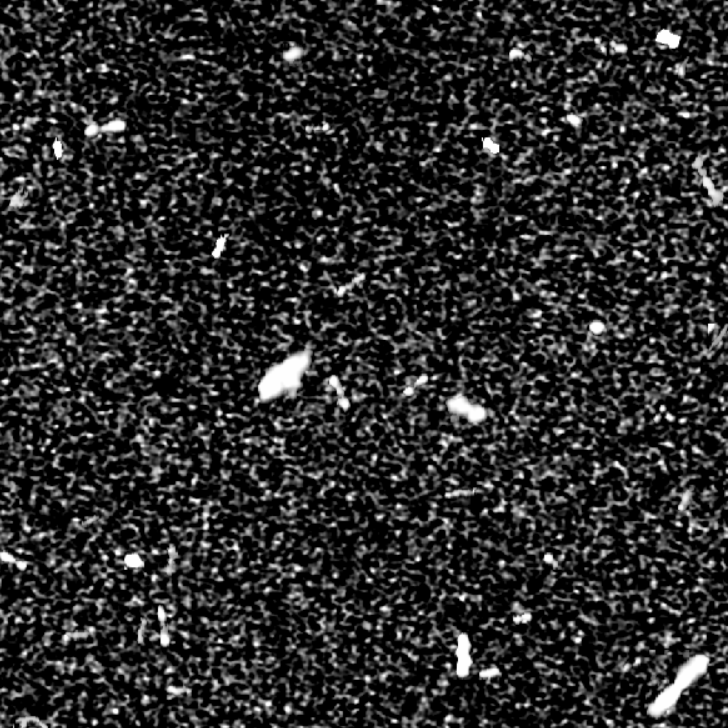}
        \label{subfig:large2}}
    \subfloat[]
        {\includegraphics[width=0.33\textwidth]{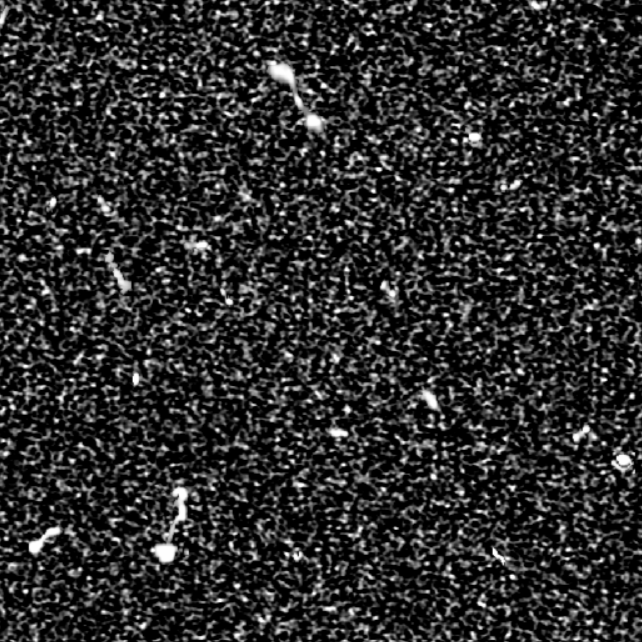}
        \label{subfig:large3}}
    \vspace{2mm}
  \caption{Crops of the large-scale map with synthetic objects on a real background noise map.}
  \label{fig:large-scale}
\end{figure}

Exploiting the capabilities of our model, able to generate high-quality samples with well-defined shapes, we additionally evaluate the RADiff pipeline in the generation of synthetic large-scale maps populated with radio-astronomical objects for Data Challenge purposes.
To achieve this, we employ our RADiff model, pre-trained on the SC dataset, to guide the image generation using $N$ semantic masks, (thus generating $N$ crops of size 128 $\times$ 128), then filter out the background from each crop, extract the objects from the generated image, and paste them on a real large background map. Since the images generated by the model contain values in the range $\left[0, 1 \right]$, we need to rescale these values to adapt them to the flux of the background noise map. We modulate the pixel values of each object by the standard deviation of the background flux $\sigma$ and by a scaling factor $k$ sampled from an exponential distribution. We choose a $\lambda$ value of 3 for the exponential distribution and limit its maximum value to 10 in the images we show in this analysis, even though these values are parameterizable.
Finally, we place the objects randomly on the background map, avoiding overlap. The distribution and the percentage of accepted overlap between the objects are also parameterizable.
Some zoomed-in visual results of this approach are shown in Figure~\ref{fig:large-scale}, while more detailed images at different zoom levels are reported in Section~\ref{sec:appendix}.
Since the model is trained by applying some pre-processing to the input images, the generated objects will contain values that follow the distribution of the processed ones. 
In the future, we will therefore investigate generative approaches with modular preprocessing functions applied, allowing users to choose the image transformations that best suit their use case.

\label{sec:ablations}

\section{Conclusion}

In this work, we leverage the flexibility of Latent Diffusion Models with conditioning information by proposing RADiff, a conditional LDM that conditions the generation of synthetic images with a semantic mask and background information. We apply this pipeline to augment small datasets, evaluating how extending a real dataset with synthetic samples impacts the performance of deep learning models. We evaluated the impact of using synthetic and generated masks to generate synthetic images and trained a semantic segmentation model on these augmented datasets. In this evaluation, we found an improvement in performance, especially when treating class imbalances.
We also exploited the quality of the generated samples to simulate large-scale radio maps populated with different kinds of objects to be used in data challenge simulations.
Further exploration of our work may include the extension of this pipeline to 3D data, enabling the generation of 3D spectral data cubes, and the improvement of large-scale map generations with better-designed post-processing operations or with architectural changes to the pipeline to better suit the task.

\label{sec:conclusions}

\section*{Acknowledgements}

Part of this work has been supported by the INAF PRIN TEC CIRASA programme, and the spokes "FutureHPC \& BigData” and "Astrophysics \& Cosmos Observations" of the ICSC – Centro Nazionale di Ricerca in High Performance Computing, Big Data and Quantum Computing – and hosting entity, funded by European Union – NextGenerationEU.
\label{sec:acknowledgements}

\appendix
\section{Additional Large Scale Samples}
\begin{figure*}
    \centering
    \includegraphics[width=0.8\textwidth]{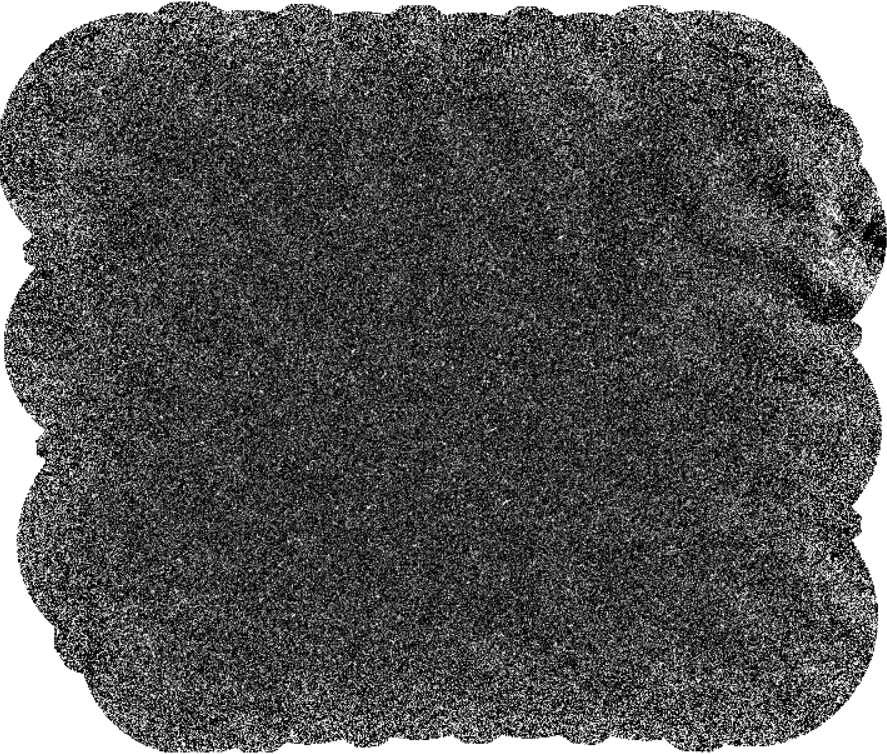}
    \caption{Large-scale background noise map populated with synthetic objects. Fully-zoomed-out image.}
    \label{fig:large-zoom-out}
\end{figure*}

\begin{figure*}
    \centering
    \includegraphics[width=0.8\textwidth]{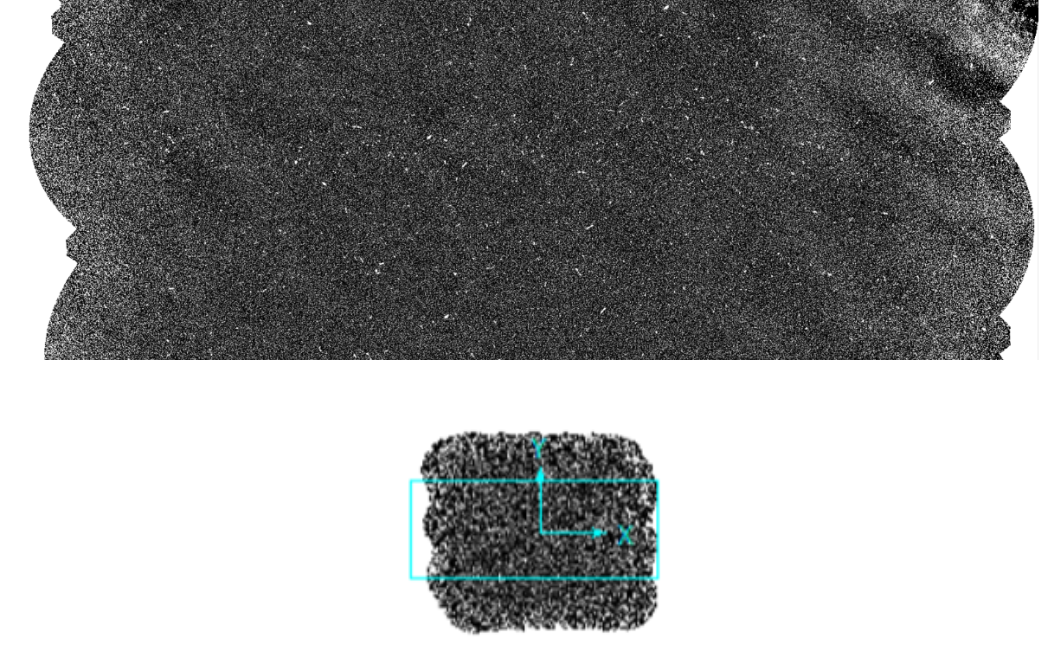}
    \caption{Large-scale background noise map populated with synthetic objects. Zoom at 2$\times$. The lower image indicates the focused region in the complete map.}
    \label{fig:large-zoom1}
\end{figure*}

\begin{figure*}
    \centering
    \includegraphics[width=0.8\textwidth]{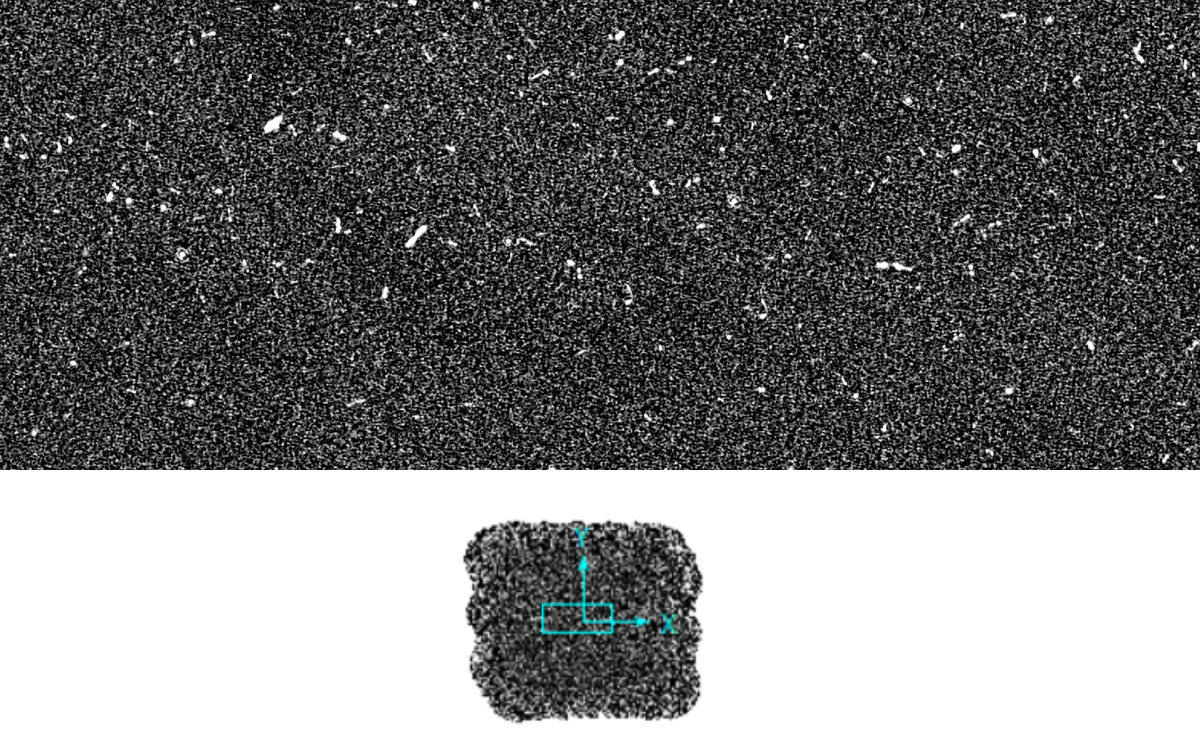}
    \caption{Large-scale background noise map populated with synthetic objects. Zoom at 5$\times$. The lower image indicates the focused region in the complete map.}
    \label{fig:large-zoom2}
\end{figure*}

\begin{figure*}
    \centering
    \includegraphics[width=0.8\textwidth]{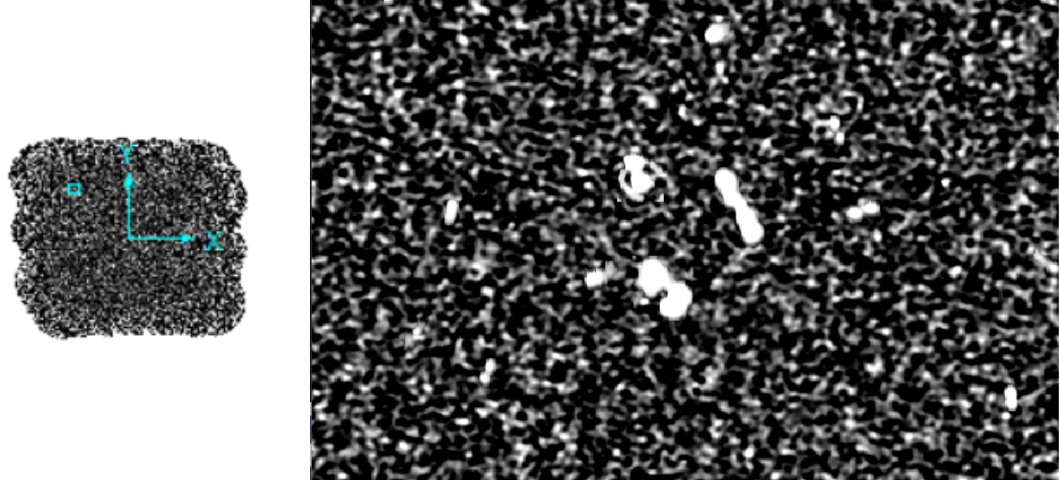}
    \caption{Large-scale background noise map populated with synthetic objects. 10$\times$ zoomed region. The focused region is indicated on the leftmost image.}
    \label{fig:large-zoom3}
\end{figure*}

\begin{figure*}
    \centering
    \includegraphics[width=0.8\textwidth]{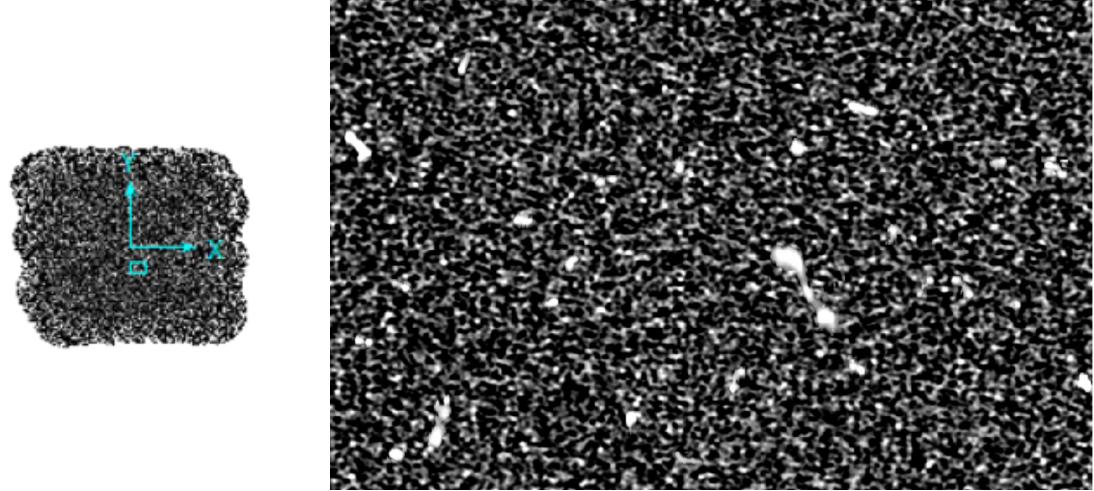}
    \caption{Large-scale background noise map populated with synthetic objects. 10$\times$ zoomed region. The focused region is indicated on the leftmost image.}
    \label{fig:large-zoom4}
\end{figure*}

In this section, we report additional samples of the large-scale generated map with the same method presented in Section~\ref{sec:large-scale}. Figure \ref{fig:large-zoom-out} reports the whole map populated with generated objects, at its total size. Since this has a high number of pixels (12,000 $\times$ 15,000), small details and the shape of the objects are not visible in this representation. For this reason, we also report the same image at several levels of zoom, indicating the focused region as well. Figure~\ref{fig:large-zoom1} shows the same image with a 2$\times$-zoom, highlighting how the objects are distributed in the map. This can be better visualized in the 5$\times$-zoomed image, reported in Figure~\ref{fig:large-zoom2}. Figures~\ref{fig:large-zoom3} and~\ref{fig:large-zoom4} show two different regions zoomed at a 10$\times$ scale, focusing on the shape of the synthetic objects.

\label{sec:appendix}

\bibliographystyle{unsrtnat}
\bibliography{cas-refs}  

\begin{thebibliography}{68}
\providecommand{\natexlab}[1]{#1}
\providecommand{\url}[1]{\texttt{#1}}
\expandafter\ifx\csname urlstyle\endcsname\relax
  \providecommand{\doi}[1]{doi: #1}\else
  \providecommand{\doi}{doi: \begingroup \urlstyle{rm}\Url}\fi

\bibitem[Dewdney et~al.(2009)Dewdney, Hall, Schilizzi, and
  Lazio]{dewdney2009square}
Peter~E Dewdney, Peter~J Hall, Richard~T Schilizzi, and T~Joseph~LW Lazio.
\newblock The square kilometre array.
\newblock \emph{Proceedings of the IEEE}, 97\penalty0 (8):\penalty0 1482--1496,
  2009.

\bibitem[Hotan et~al.(2021)Hotan, Bunton, Chippendale, Whiting, Tuthill, Moss,
  McConnell, Amy, Huynh, Allison, and et~al.]{hotan2021}
A.~W. Hotan, J.~D. Bunton, A.~P. Chippendale, M.~Whiting, J.~Tuthill, V.~A.
  Moss, D.~McConnell, S.~W. Amy, M.~T. Huynh, J.~R. Allison, and et~al.
\newblock Australian square kilometre array pathfinder: I. system description.
\newblock \emph{Publications of the Astronomical Society of Australia},
  38:\penalty0 e009, 2021.
\newblock \doi{10.1017/pasa.2021.1}.

\bibitem[Sortino et~al.(2023)Sortino, Magro, Fiameni, Sciacca, Riggi, DeMarco,
  Spampinato, Hopkins, Bufano, Schillir{\`o}, et~al.]{sortino2023radio}
Renato Sortino, Daniel Magro, Giuseppe Fiameni, Eva Sciacca, Simone Riggi,
  Andrea DeMarco, Concetto Spampinato, Andrew~M Hopkins, Filomena Bufano,
  Francesco Schillir{\`o}, et~al.
\newblock Radio astronomical images object detection and segmentation: a
  benchmark on deep learning methods.
\newblock \emph{Experimental Astronomy}, pages 1--39, 2023.

\bibitem[Wu et~al.(2019)Wu, Wong, Rudnick, Shabala, Alger, Banfield, Ong,
  White, Garon, Norris, Andernach, Tate, Lukic, Tang, Schawinski, and
  Diakogiannis]{wu2019radio}
Chen Wu, Oiwei~Ivy Wong, Lawrence Rudnick, Stanislav~S Shabala, Matthew~J
  Alger, Julie~K Banfield, Cheng~Soon Ong, Sarah~V White, Avery~F Garon, Ray~P
  Norris, Heinz Andernach, Jean Tate, Vesna Lukic, Hongming Tang, Kevin
  Schawinski, and Foivos~I Diakogiannis.
\newblock Radio galaxy zoo: Claran – a deep learning classifier for radio
  morphologies.
\newblock \emph{Monthly Notices of the Royal Astronomical Society},
  482\penalty0 (1):\penalty0 1211--1230, 2019.
\newblock \doi{10.1093/mnras/sty2646}.
\newblock URL \url{http://dx.doi.org/10.1093/mnras/sty2646}.

\bibitem[Gheller et~al.(2018)Gheller, Vazza, and Bonafede]{gheller2018deep}
Claudio Gheller, Franco Vazza, and Annalisa Bonafede.
\newblock Deep learning based detection of cosmological diffuse radio sources.
\newblock \emph{Monthly Notices of the Royal Astronomical Society},
  480\penalty0 (3):\penalty0 3749--3761, 2018.

\bibitem[Becker et~al.(2021)Becker, Vaccari, Prescott, and
  Grobler]{becker2021cnn}
Burger Becker, Mattia Vaccari, Matthew Prescott, and Trienko Grobler.
\newblock Cnn architecture comparison for radio galaxy classification.
\newblock \emph{Monthly Notices of the Royal Astronomical Society},
  503\penalty0 (2):\penalty0 1828--1846, 2021.

\bibitem[Lukic et~al.(2019)Lukic, de~Gasperin, and
  Br{\"u}ggen]{lukic2019convosource}
Vesna Lukic, Francesco de~Gasperin, and Marcus Br{\"u}ggen.
\newblock Convosource: radio-astronomical source-finding with convolutional
  neural networks.
\newblock \emph{Galaxies}, 8\penalty0 (1):\penalty0 3, 2019.

\bibitem[Bonaldi et~al.(2021)]{bonaldi2021square}
A~Bonaldi et~al.
\newblock Square kilometre array science data challenge 1: analysis and
  results.
\newblock \emph{Monthly Notices of the Royal Astronomical Society},
  500\penalty0 (3):\penalty0 3821--3837, 2021.

\bibitem[Hopkins et~al.(2015)Hopkins, Whiting, Seymour, Chow, Norris, Bonavera,
  Breton, Carbone, Ferrari, Franzen, et~al.]{hopkins2015askap}
Andrew~M Hopkins, Matthew~T Whiting, Nick Seymour, KE~Chow, Ray~P Norris, Laura
  Bonavera, Rene Breton, Dario Carbone, Chiara Ferrari, TMO Franzen, et~al.
\newblock The askap/emu source finding data challenge.
\newblock \emph{Publications of the Astronomical Society of Australia},
  32:\penalty0 e037, 2015.

\bibitem[Riggi et~al.(2019)Riggi, Vitello, Becciani, Buemi, Bufano, Calanducci,
  Cavallaro, Costa, Ingallinera, Leto, et~al.]{riggi2019caesar}
Simone Riggi, F~Vitello, Ugo Becciani, C~Buemi, FILOMENA Bufano, A~Calanducci,
  Francesco Cavallaro, Alessandro Costa, Adriano Ingallinera, PAOLO Leto,
  et~al.
\newblock Caesar source finder: Recent developments and testing.
\newblock \emph{Publications of the Astronomical Society of Australia}, 36,
  2019.

\bibitem[Boyce et~al.(2023)Boyce, Hopkins, Riggi, Rudnick, Ramsay, Hale,
  Marvil, Whiting, Venkataraman, O'Dea, et~al.]{boyce2023hydra}
MM~Boyce, AM~Hopkins, S~Riggi, L~Rudnick, M~Ramsay, CL~Hale, J~Marvil,
  M~Whiting, P~Venkataraman, CP~O'Dea, et~al.
\newblock Hydra ii: Characterisation of aegean, caesar, profound, pybdsf, and
  selavy source finders.
\newblock \emph{arXiv preprint arXiv:2304.14357}, 2023.

\bibitem[Kucera et~al.(2022)Kucera, Togninalli, and
  Meng-Papaxanthos]{kucera2022conditional}
Tim Kucera, Matteo Togninalli, and Laetitia Meng-Papaxanthos.
\newblock Conditional generative modeling for de novo protein design with
  hierarchical functions.
\newblock \emph{Bioinformatics}, 38\penalty0 (13):\penalty0 3454--3461, 2022.

\bibitem[Havaei et~al.(2021)Havaei, Mao, Wang, and Lao]{havaei2021conditional}
Mohammad Havaei, Ximeng Mao, Yiping Wang, and Qicheng Lao.
\newblock Conditional generation of medical images via disentangled adversarial
  inference.
\newblock \emph{Medical Image Analysis}, 72:\penalty0 102106, 2021.

\bibitem[Antoniou et~al.(2017)Antoniou, Storkey, and Edwards]{antoniou2017data}
Antreas Antoniou, Amos Storkey, and Harrison Edwards.
\newblock Data augmentation generative adversarial networks.
\newblock \emph{arXiv preprint arXiv:1711.04340}, 2017.

\bibitem[Yin et~al.(2023)Yin, Kaddour, Zhang, Nie, Liu, Kong, and
  Liu]{yin2023ttida}
Yuwei Yin, Jean Kaddour, Xiang Zhang, Yixin Nie, Zhenguang Liu, Lingpeng Kong,
  and Qi~Liu.
\newblock Ttida: Controllable generative data augmentation via text-to-text and
  text-to-image models.
\newblock \emph{arXiv preprint arXiv:2304.08821}, 2023.

\bibitem[Sampath et~al.(2021)Sampath, Maurtua, Aguilar~Martin, and
  Gutierrez]{sampath2021survey}
Vignesh Sampath, I{\~n}aki Maurtua, Juan~Jose Aguilar~Martin, and Aitor
  Gutierrez.
\newblock A survey on generative adversarial networks for imbalance problems in
  computer vision tasks.
\newblock \emph{Journal of big Data}, 8:\penalty0 1--59, 2021.

\bibitem[Goodfellow et~al.(2014)Goodfellow, Pouget-Abadie, Mirza, Xu,
  Warde-Farley, Ozair, Courville, and Bengio]{goodfellow2014generative}
Ian Goodfellow, Jean Pouget-Abadie, Mehdi Mirza, Bing Xu, David Warde-Farley,
  Sherjil Ozair, Aaron Courville, and Yoshua Bengio.
\newblock Generative adversarial nets.
\newblock In Z.~Ghahramani, M.~Welling, C.~Cortes, N.~Lawrence, and K.Q.
  Weinberger, editors, \emph{Advances in Neural Information Processing
  Systems}, volume~27. Curran Associates, Inc., 2014.
\newblock URL
  \url{https://proceedings.neurips.cc/paper_files/paper/2014/file/5ca3e9b122f61f8f06494c97b1afccf3-Paper.pdf}.

\bibitem[Glaser et~al.(2019)Glaser, Wong, Schawinski, and
  Zhang]{glaser2019radiogan}
Nina Glaser, O~Ivy Wong, Kevin Schawinski, and Ce~Zhang.
\newblock Radiogan--translations between different radio surveys with
  generative adversarial networks.
\newblock \emph{Monthly Notices of the Royal Astronomical Society},
  487\penalty0 (3):\penalty0 4190--4207, 2019.

\bibitem[Balakrishnan et~al.(2021)Balakrishnan, Champion, Barr, Kramer, Sengar,
  and Bailes]{balakrishnan2021pulsar}
Vishnu Balakrishnan, David Champion, Ewan Barr, Michael Kramer, Rahul Sengar,
  and Matthew Bailes.
\newblock Pulsar candidate identification using semi-supervised generative
  adversarial networks.
\newblock \emph{Monthly Notices of the Royal Astronomical Society},
  505\penalty0 (1):\penalty0 1180--1194, 2021.

\bibitem[Guo et~al.(2019)Guo, Duan, Wang, Yao, Yin, Xin, Li, Qian, Wang, Pan,
  et~al.]{guo2019pulsar}
Ping Guo, Fuqing Duan, Pei Wang, Yao Yao, Qian Yin, Xin Xin, Di~Li, Lei Qian,
  Shen Wang, Zhichen Pan, et~al.
\newblock Pulsar candidate classification using generative adversary networks.
\newblock \emph{Monthly Notices of the Royal Astronomical Society},
  490\penalty0 (4):\penalty0 5424--5439, 2019.

\bibitem[Ho et~al.(2020)Ho, Jain, and Abbeel]{ho2020denoising}
Jonathan Ho, Ajay Jain, and Pieter Abbeel.
\newblock Denoising diffusion probabilistic models.
\newblock \emph{Advances in Neural Information Processing Systems},
  33:\penalty0 6840--6851, 2020.

\bibitem[Dhariwal and Nichol(2021)]{dhariwal2021diffusion}
Prafulla Dhariwal and Alexander Nichol.
\newblock Diffusion models beat gans on image synthesis.
\newblock \emph{Advances in Neural Information Processing Systems},
  34:\penalty0 8780--8794, 2021.

\bibitem[Rombach et~al.(2022)Rombach, Blattmann, Lorenz, Esser, and
  Ommer]{rombach2022high}
Robin Rombach, Andreas Blattmann, Dominik Lorenz, Patrick Esser, and Bj{\"o}rn
  Ommer.
\newblock High-resolution image synthesis with latent diffusion models.
\newblock In \emph{Proceedings of the IEEE/CVF Conference on Computer Vision
  and Pattern Recognition}, pages 10684--10695, 2022.

\bibitem[Huang et~al.(2023)Huang, Chen, Liu, Shen, Zhao, and
  Zhou]{huang2023composer}
Lianghua Huang, Di~Chen, Yu~Liu, Yujun Shen, Deli Zhao, and Jingren Zhou.
\newblock Composer: Creative and controllable image synthesis with composable
  conditions.
\newblock \emph{arXiv preprint arXiv:2302.09778}, 2023.

\bibitem[Zhang and Agrawala(2023)]{zhang2023adding}
Lvmin Zhang and Maneesh Agrawala.
\newblock Adding conditional control to text-to-image diffusion models.
\newblock \emph{arXiv preprint arXiv:2302.05543}, 2023.

\bibitem[Wang et~al.(2023)Wang, Chen, Luo, and Wang]{wang2023conditional}
Ruoqi Wang, Zhuoyang Chen, Qiong Luo, and Feng Wang.
\newblock A conditional denoising diffusion probabilistic model for radio
  interferometric image reconstruction.
\newblock \emph{arXiv preprint arXiv:2305.09121}, 2023.

\bibitem[Kingma and Welling(2013)]{kingma2013auto}
Diederik~P Kingma and Max Welling.
\newblock Auto-encoding variational bayes.
\newblock \emph{arXiv preprint arXiv:1312.6114}, 2013.

\bibitem[van~den Oord et~al.(2017)van~den Oord, Vinyals, and
  kavukcuoglu]{van2017neural}
Aaron van~den Oord, Oriol Vinyals, and koray kavukcuoglu.
\newblock Neural discrete representation learning.
\newblock In I.~Guyon, U.~Von Luxburg, S.~Bengio, H.~Wallach, R.~Fergus,
  S.~Vishwanathan, and R.~Garnett, editors, \emph{Advances in Neural
  Information Processing Systems}, volume~30. Curran Associates, Inc., 2017.
\newblock URL
  \url{https://proceedings.neurips.cc/paper_files/paper/2017/file/7a98af17e63a0ac09ce2e96d03992fbc-Paper.pdf}.

\bibitem[Zhao et~al.(2017)Zhao, Zhao, and Eskenazi]{zhao2017learning}
Tiancheng Zhao, Ran Zhao, and Maxine Eskenazi.
\newblock Learning discourse-level diversity for neural dialog models using
  conditional variational autoencoders.
\newblock \emph{arXiv preprint arXiv:1703.10960}, 2017.

\bibitem[Lin et~al.(2019)Lin, Liu, and Liang]{lin2019improving}
Lizhi Lin, Xinyue Liu, and Wenxin Liang.
\newblock Improving variational auto-encoder with self-attention and mutual
  information for image generation.
\newblock In \emph{Proceedings of the 3rd international conference on video and
  image processing}, pages 162--167, 2019.

\bibitem[Mirza and Osindero(2014)]{mirza2014conditional}
Mehdi Mirza and Simon Osindero.
\newblock Conditional generative adversarial nets.
\newblock \emph{arXiv preprint arXiv:1411.1784}, 2014.

\bibitem[Isola et~al.(2017)Isola, Zhu, Zhou, and Efros]{isola2017image}
Phillip Isola, Jun-Yan Zhu, Tinghui Zhou, and Alexei~A Efros.
\newblock Image-to-image translation with conditional adversarial networks.
\newblock In \emph{Proceedings of the IEEE conference on computer vision and
  pattern recognition}, pages 1125--1134, 2017.

\bibitem[Reed et~al.(2016)Reed, Akata, Yan, Logeswaran, Schiele, and
  Lee]{reed2016generative}
Scott Reed, Zeynep Akata, Xinchen Yan, Lajanugen Logeswaran, Bernt Schiele, and
  Honglak Lee.
\newblock Generative adversarial text to image synthesis.
\newblock In \emph{International conference on machine learning}, pages
  1060--1069. PMLR, 2016.

\bibitem[Karras et~al.(2020)Karras, Laine, Aittala, Hellsten, Lehtinen, and
  Aila]{karras2020analyzing}
Tero Karras, Samuli Laine, Miika Aittala, Janne Hellsten, Jaakko Lehtinen, and
  Timo Aila.
\newblock Analyzing and improving the image quality of stylegan.
\newblock In \emph{Proceedings of the IEEE/CVF conference on computer vision
  and pattern recognition}, pages 8110--8119, 2020.

\bibitem[Zhu et~al.(2017)Zhu, Park, Isola, and Efros]{zhu2017unpaired}
Jun-Yan Zhu, Taesung Park, Phillip Isola, and Alexei~A Efros.
\newblock Unpaired image-to-image translation using cycle-consistent
  adversarial networks.
\newblock In \emph{Proceedings of the IEEE international conference on computer
  vision}, pages 2223--2232, 2017.

\bibitem[Bastien et~al.(2021)Bastien, Scaife, Tang, Bowles, and
  Porter]{bastien2021structured}
David~J Bastien, Anna~MM Scaife, Hongming Tang, Micah Bowles, and Fiona Porter.
\newblock Structured variational inference for simulating populations of radio
  galaxies.
\newblock \emph{Monthly Notices of the Royal Astronomical Society},
  503\penalty0 (3):\penalty0 3351--3370, 2021.

\bibitem[Spindler et~al.(2021)Spindler, Geach, and
  Smith]{spindler2021astrovader}
Ashley Spindler, James~E Geach, and Michael~J Smith.
\newblock Astrovader: astronomical variational deep embedder for unsupervised
  morphological classification of galaxies and synthetic image generation.
\newblock \emph{Monthly Notices of the Royal Astronomical Society},
  502\penalty0 (1):\penalty0 985--1007, 2021.

\bibitem[Lanusse et~al.(2021)Lanusse, Mandelbaum, Ravanbakhsh, Li, Freeman, and
  Póczos]{Lanusse2021}
François Lanusse, Rachel Mandelbaum, Siamak Ravanbakhsh, Chun-Liang Li, Peter
  Freeman, and Barnabás Póczos.
\newblock {Deep generative models for galaxy image simulations}.
\newblock \emph{Monthly Notices of the Royal Astronomical Society},
  504\penalty0 (4):\penalty0 5543--5555, 05 2021.
\newblock ISSN 0035-8711.
\newblock \doi{10.1093/mnras/stab1214}.
\newblock URL \url{https://doi.org/10.1093/mnras/stab1214}.

\bibitem[Scully et~al.(2023)Scully, Flynn, Carley, Gallagher, and
  Daly]{scully2023simulating}
Jeremiah Scully, Ronan Flynn, Eoin Carley, Peter Gallagher, and Mark Daly.
\newblock Simulating solar radio bursts using generative adversarial networks.
\newblock \emph{Solar Physics}, 298\penalty0 (1):\penalty0 1--16, 2023.

\bibitem[Kummer et~al.(2022)Kummer, Rustige, Griese, Borras, Br{\"u}ggen,
  Connor, Gaede, Kasieczka, and Schleper]{kummer2022radio}
Janis Kummer, Lennart Rustige, Florian Griese, Kerstin Borras, Marcus
  Br{\"u}ggen, Patrick~LS Connor, Frank Gaede, Gregor Kasieczka, and Peter
  Schleper.
\newblock Radio galaxy classification with wgan-supported augmentation.
\newblock \emph{arXiv preprint arXiv:2206.15131}, 2022.

\bibitem[Umana and et~al.(2015)]{Umana2015}
Grazia Umana and et~al.
\newblock Scorpio: a deep survey of radio emission from the stellar life-cycle.
\newblock \emph{MNRAS}, 454:\penalty0 902--912, 2015.

\bibitem[Becker et~al.(1995)Becker, White, and Helfand]{becker1995first}
Robert~H Becker, Richard~L White, and David~J Helfand.
\newblock The first survey: faint images of the radio sky at twenty
  centimeters.
\newblock \emph{The Astrophysical Journal}, 450:\penalty0 559, 1995.

\bibitem[Banfield et~al.(2015)Banfield, Wong, Willett, Norris, Rudnick,
  Shabala, Simmons, Snyder, Garon, Seymour, et~al.]{Banfield2015}
Julie~K Banfield, OI~Wong, Kylie~W Willett, Richard~P Norris, L~Rudnick,
  Stanislav~S Shabala, Brooke~D Simmons, Chris Snyder, A~Garon, Nick Seymour,
  et~al.
\newblock Radio galaxy zoo: host galaxies and radio morphologies derived from
  visual inspection.
\newblock \emph{Monthly Notices of the Royal Astronomical Society},
  453\penalty0 (3):\penalty0 2326--2340, 2015.

\bibitem[Ingallinera et~al.(2022)Ingallinera, Cavallaro, Loru, Marvil, Umana,
  Trigilio, Breen, Bordiu, Buemi, Bufano, et~al.]{ingallinera2022evolutionary}
Adriano Ingallinera, Francesco Cavallaro, Sara Loru, J~Marvil, G~Umana, CORRADO
  Trigilio, S~Breen, C~Bordiu, CARLA~SIMONA Buemi, FILOMENA Bufano, et~al.
\newblock Evolutionary map of the universe (emu): 18-cm oh-maser discovery in
  askap continuum images of the scorpio field.
\newblock \emph{Monthly Notices of the Royal Astronomical Society: Letters},
  512\penalty0 (1):\penalty0 L21--L26, 2022.

\bibitem[Umana et~al.(2021)Umana, Trigilio, Ingallinera, Riggi, Cavallaro,
  Marvil, Norris, Hopkins, Buemi, Bufano, Leto, Loru, Bordiu, Bunton, Collier,
  Filipovic, Franzen, Thompson, Andernach, Carretti, Dai, Kapińska,
  Koribalski, Kothes, Leahy, Mcconnell, Tothill, and
  Michałowski]{umana2021first}
G~Umana, C~Trigilio, A~Ingallinera, S~Riggi, F~Cavallaro, J~Marvil, R~P Norris,
  A~M Hopkins, C~S Buemi, F~Bufano, P~Leto, S~Loru, C~Bordiu, J~D Bunton, J~D
  Collier, M~Filipovic, T~M~O Franzen, M~A Thompson, H~Andernach, E~Carretti,
  S~Dai, A~Kapińska, B~S Koribalski, R~Kothes, D~Leahy, D~Mcconnell,
  N~Tothill, and M~J Michałowski.
\newblock {A first glimpse at the Galactic plane with the ASKAP: the SCORPIO
  field}.
\newblock \emph{Monthly Notices of the Royal Astronomical Society},
  506\penalty0 (2):\penalty0 2232--2246, 05 2021.
\newblock ISSN 0035-8711.
\newblock \doi{10.1093/mnras/stab1279}.
\newblock URL \url{https://doi.org/10.1093/mnras/stab1279}.

\bibitem[Norris et~al.(2021)Norris, Marvil, Collier, Kapi{\'n}ska, O’Brien,
  Rudnick, Andernach, Asorey, Brown, Br{\"u}ggen, et~al.]{Norris2021}
Ray~P Norris, Joshua Marvil, Jordan~D Collier, Anna~D Kapi{\'n}ska, Andrew~N
  O’Brien, L~Rudnick, Heinz Andernach, Jacobo Asorey, Michael~JI Brown,
  Marcus Br{\"u}ggen, et~al.
\newblock The evolutionary map of the universe pilot survey.
\newblock \emph{Publications of the Astronomical Society of Australia}, 38,
  2021.

\bibitem[Riggi et~al.(2023)Riggi, Magro, Sortino, De~Marco, Bordiu, Cecconello,
  Hopkins, Marvil, Umana, Sciacca, et~al.]{riggi2023astronomical}
S~Riggi, D~Magro, R~Sortino, A~De~Marco, C~Bordiu, T~Cecconello, AM~Hopkins,
  J~Marvil, G~Umana, E~Sciacca, et~al.
\newblock Astronomical source detection in radio continuum maps with deep
  neural networks.
\newblock \emph{Astronomy and Computing}, 42:\penalty0 100682, 2023.

\bibitem[Ronneberger et~al.(2015)Ronneberger, Fischer, and
  Brox]{ronneberger2015unet}
Olaf Ronneberger, Philipp Fischer, and Thomas Brox.
\newblock U-net: Convolutional networks for biomedical image segmentation.
\newblock In \emph{International Conference on Medical image computing and
  computer-assisted intervention}, pages 234--241. Springer, 2015.

\bibitem[Pino et~al.(2021)]{pino2021semantic}
Carmelo Pino et~al.
\newblock Semantic segmentation of radio-astronomical images.
\newblock In \emph{International Workshop on Artificial Intelligence and
  Pattern Recognition}, pages 393--403. Springer, 2021.

\bibitem[Xie et~al.(2021)Xie, Wang, Yu, Anandkumar, Alvarez, and
  Luo]{xie2021segformer}
Enze Xie, Wenhai Wang, Zhiding Yu, Anima Anandkumar, Jose~M Alvarez, and Ping
  Luo.
\newblock Segformer: Simple and efficient design for semantic segmentation with
  transformers.
\newblock \emph{Advances in Neural Information Processing Systems},
  34:\penalty0 12077--12090, 2021.

\bibitem[Vaswani et~al.(2017)Vaswani, Shazeer, Parmar, Uszkoreit, Jones, Gomez,
  Kaiser, and Polosukhin]{vaswani2017attention}
Ashish Vaswani, Noam Shazeer, Niki Parmar, Jakob Uszkoreit, Llion Jones,
  Aidan~N Gomez, {\L}ukasz Kaiser, and Illia Polosukhin.
\newblock Attention is all you need.
\newblock \emph{Advances in neural information processing systems}, 30, 2017.

\bibitem[Hinton and Salakhutdinov(2006)]{hinton2006reducing}
G.~E. Hinton and R.~R. Salakhutdinov.
\newblock Reducing the dimensionality of data with neural networks.
\newblock \emph{Science}, 313\penalty0 (5786):\penalty0 504--507, 2006.
\newblock \doi{10.1126/science.1127647}.
\newblock URL \url{https://www.science.org/doi/abs/10.1126/science.1127647}.

\bibitem[He et~al.(2015)He, Zhang, Ren, and Sun]{he2014deep}
Kaiming He, Xiangyu Zhang, Shaoqing Ren, and Jian Sun.
\newblock Deep residual learning for image recognition. arxiv 2015.
\newblock \emph{arXiv preprint arXiv:1512.03385}, 14, 2015.

\bibitem[Heusel et~al.(2017)]{heusel2017gans}
Martin Heusel et~al.
\newblock Gans trained by a two time-scale update rule converge to a local nash
  equilibrium.
\newblock \emph{Advances in neural information processing systems}, 30, 2017.

\bibitem[Szegedy et~al.(2016)]{szegedy2016rethinking}
Christian Szegedy et~al.
\newblock Rethinking the inception architecture for computer vision.
\newblock In \emph{Proceedings of the IEEE conference on computer vision and
  pattern recognition}, pages 2818--2826, 2016.

\bibitem[Deng et~al.(2009)Deng, Dong, Socher, Li, Li, and
  Fei-Fei]{deng2009imagenet}
Jia Deng, Wei Dong, Richard Socher, Li-Jia Li, Kai Li, and Li~Fei-Fei.
\newblock Imagenet: A large-scale hierarchical image database.
\newblock In \emph{2009 IEEE conference on computer vision and pattern
  recognition}, pages 248--255. Ieee, 2009.

\bibitem[Morozov et~al.(2021)Morozov, Voynov, and Babenko]{morozov2021self}
Stanislav Morozov, Andrey Voynov, and Artem Babenko.
\newblock On self-supervised image representations for gan evaluation.
\newblock In \emph{International Conference on Learning Representations}, 2021.

\bibitem[Grill et~al.(2020)Grill, Strub, Altch{\'e}, Tallec, Richemond,
  Buchatskaya, Doersch, Avila~Pires, Guo, Gheshlaghi~Azar,
  et~al.]{grill2020bootstrap}
Jean-Bastien Grill, Florian Strub, Florent Altch{\'e}, Corentin Tallec, Pierre
  Richemond, Elena Buchatskaya, Carl Doersch, Bernardo Avila~Pires, Zhaohan
  Guo, Mohammad Gheshlaghi~Azar, et~al.
\newblock Bootstrap your own latent-a new approach to self-supervised learning.
\newblock \emph{Advances in neural information processing systems},
  33:\penalty0 21271--21284, 2020.

\bibitem[Radford et~al.(2015)Radford, Metz, and
  Chintala]{radford2015unsupervised}
Alec Radford, Luke Metz, and Soumith Chintala.
\newblock Unsupervised representation learning with deep convolutional
  generative adversarial networks.
\newblock \emph{arXiv preprint arXiv:1511.06434}, 2015.

\bibitem[Karras et~al.(2017)]{karras2017progressive}
Tero Karras et~al.
\newblock Progressive growing of gans for improved quality, stability, and
  variation.
\newblock \emph{arXiv preprint arXiv:1710.10196}, 2017.

\bibitem[Zhang et~al.(2022)Zhang, Wu, Zhang, Zhu, Lin, Zhang, Sun, He, Mueller,
  Manmatha, et~al.]{zhang2022resnest}
Hang Zhang, Chongruo Wu, Zhongyue Zhang, Yi~Zhu, Haibin Lin, Zhi Zhang, Yue
  Sun, Tong He, Jonas Mueller, R~Manmatha, et~al.
\newblock Resnest: Split-attention networks.
\newblock In \emph{Proceedings of the IEEE/CVF Conference on Computer Vision
  and Pattern Recognition}, pages 2736--2746, 2022.

\bibitem[Cheng et~al.(2020)Cheng, Collins, Zhu, Liu, Huang, Adam, and
  Chen]{cheng2020panoptic}
Bowen Cheng, Maxwell~D Collins, Yukun Zhu, Ting Liu, Thomas~S Huang, Hartwig
  Adam, and Liang-Chieh Chen.
\newblock Panoptic-deeplab: A simple, strong, and fast baseline for bottom-up
  panoptic segmentation.
\newblock In \emph{Proceedings of the IEEE/CVF conference on computer vision
  and pattern recognition}, pages 12475--12485, 2020.

\bibitem[Bao et~al.(2021)Bao, Dong, Piao, and Wei]{bao2021beit}
Hangbo Bao, Li~Dong, Songhao Piao, and Furu Wei.
\newblock Beit: Bert pre-training of image transformers.
\newblock \emph{arXiv preprint arXiv:2106.08254}, 2021.

\bibitem[J{\'e}gou et~al.(2017)]{jegou2017one}
Simon J{\'e}gou et~al.
\newblock The one hundred layers tiramisu: Fully convolutional densenets for
  semantic segmentation.
\newblock In \emph{Proceedings of the IEEE conference on computer vision and
  pattern recognition workshops}, pages 11--19, 2017.

\bibitem[Wang et~al.(2004)Wang, Bovik, Sheikh, and Simoncelli]{wang2004image}
Zhou Wang, Alan~C Bovik, Hamid~R Sheikh, and Eero~P Simoncelli.
\newblock Image quality assessment: from error visibility to structural
  similarity.
\newblock \emph{IEEE Trans. Image Processing}, 13\penalty0 (4):\penalty0
  600--612, 2004.

\bibitem[Park et~al.(2019)Park, Liu, Wang, and Zhu]{park2019semantic}
Taesung Park, Ming-Yu Liu, Ting-Chun Wang, and Jun-Yan Zhu.
\newblock Semantic image synthesis with spatially-adaptive normalization.
\newblock In \emph{Proceedings of the IEEE/CVF conference on computer vision
  and pattern recognition}, pages 2337--2346, 2019.

\bibitem[Tan et~al.(2021)Tan, Chai, Chen, Liao, Chu, Liu, Hua, and
  Yu]{tan2021diverse}
Zhentao Tan, Menglei Chai, Dongdong Chen, Jing Liao, Qi~Chu, Bin Liu, Gang Hua,
  and Nenghai Yu.
\newblock Diverse semantic image synthesis via probability distribution
  modeling.
\newblock In \emph{Proceedings of the IEEE/CVF Conference on Computer Vision
  and Pattern Recognition}, pages 7962--7971, 2021.

\bibitem[Sandfort et~al.(2019)Sandfort, Yan, Pickhardt, and
  Summers]{sandfort2019data}
Veit Sandfort, Ke~Yan, Perry~J Pickhardt, and Ronald~M Summers.
\newblock Data augmentation using generative adversarial networks (cyclegan) to
  improve generalizability in ct segmentation tasks.
\newblock \emph{Scientific reports}, 9\penalty0 (1):\penalty0 16884, 2019.

\end{thebibliography}

\end{document}